%% file: main.tex
\documentclass[10pt,twocolumn,letterpaper]{article}
\usepackage[pagenumbers]{cvpr} % To force page numbers, e.g. for an arXiv version
\usepackage{url}
\usepackage{algorithm}
\usepackage[noend]{algpseudocode}
\usepackage{amsmath}
\usepackage{dblfloatfix}
\usepackage{amssymb}
\usepackage{xspace}
\usepackage{graphicx} 
\usepackage{booktabs}
\usepackage{xcolor}         % colors
\usepackage{mathtools}      % extends amsmath
\usepackage{subcaption}     % for subfigures
\usepackage{enumitem}
\usepackage{xspace}
\usepackage{multirow}
\usepackage{wrapfig}
\usepackage[subtle]{savetrees} 
\usepackage{longtable}
\usepackage{cuted}
\usepackage{capt-of} % For \captionof
\def\setstretch#1{\renewcommand{\baselinestretch}{#1}} 
\makeatletter
\def\abstract{%
  \iftoggle{cvprpagenumbers}{}{\thispagestyle{empty}}%
  \centerline{\large\bf Abstract}%
  \vspace*{6pt}\noindent%
  \it\ignorespaces%
}

\makeatother

\makeatletter
\def\@maketitle{%
   \newpage
   \null
   \iftoggle{cvprrebuttal}{\vspace*{-.3in}}{\vskip .25in}% was .375in
   \begin{center}
      \iftoggle{cvprrebuttal}{{\large \bf \@title \par}}{{\Large \bf \@title \par}}%
      \iftoggle{cvprrebuttal}{\vspace*{-22pt}}{\vspace*{12pt}}{% was 24pt
        \large
        \lineskip .5em
        \begin{tabular}[t]{c}
          \iftoggle{cvprfinal}{\@author}{%
            \iftoggle{cvprrebuttal}{}{%
              Anonymous \confName~submission\\
              \vspace*{0pt}\\
              Paper ID \paperID
            }%
          }%
        \end{tabular}
        \par
      }%
      \vskip .5em
      \vspace*{6pt}% was 12pt
   \end{center}
}
\makeatother

\input{preamble}
\definecolor{cardinalred}{rgb}{0.549,0.082,0.082}
\definecolor{digitalred}{rgb}{0.694,0.016,0.055}

\definecolor{cvprblue}{rgb}{0.21,0.49,0.74}
\usepackage[pagebackref,breaklinks,colorlinks,allcolors=cardinalred]{hyperref}
\usepackage[capitalize]{cleveref} % for cross-referencing

\crefname{section}{Sec.}{Secs.}
\Crefname{section}{Section}{Sections}
\Crefname{table}{Table}{Tables}
\crefname{table}{Table}{Tables}
\crefformat{section}{\S#2#1#3}
\crefformat{subsection}{\S#2#1#3}
\crefformat{subsubsection}{\S#2#1#3}

\def\paperID{24462} % *** Enter the Paper ID here
\def\confName{CVPR}
\def\confYear{2026}

\title{\Large\vspace{-24pt}Scaling Verification Can Be More Effective than Scaling Policy Learning\\for Vision-Language-Action Alignment}
\author{
  Jacky Kwok\textsuperscript{1,†} \quad
  Xilun Zhang\textsuperscript{1,†} \quad
  Mengdi Xu\textsuperscript{1} \quad
  Yuejiang Liu\textsuperscript{1,§} \\[0.12cm]
  Azalia Mirhoseini\textsuperscript{1,§} \quad
  Chelsea Finn\textsuperscript{1,§} \quad
  Marco Pavone\textsuperscript{1,2,§} \\[0.2cm]
  \textsuperscript{1}Stanford University \quad \textsuperscript{2}NVIDIA Research
}

\vspace{0.06in}

\makeatletter
\let\@oldmaketitle\@maketitle
\renewcommand{\@maketitle}{
  \@oldmaketitle
  \begin{center}
    \vspace{-0.2in}
    \large
    \href{https://cover-vla.github.io}{\textcolor{digitalred}{https://cover-vla.github.io}}
    \vspace{0.15in}
    
    \includegraphics[width=\textwidth]{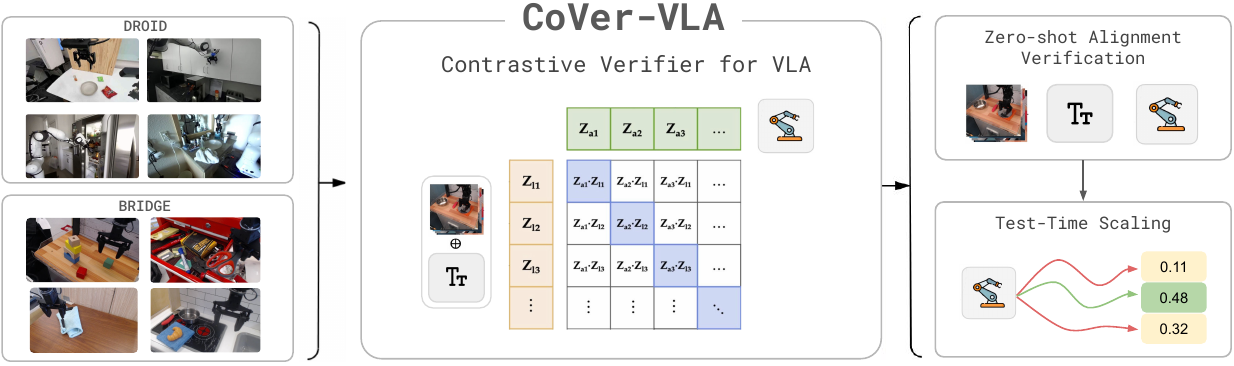}
    \captionof{figure}{We present \textbf{CoVer-VLA}, a contrastive verification framework for vision–language–action alignment.
    Our method trains a verifier on large-scale robotics datasets with contrastive learning, enabling zero-shot alignment verification for generalist robot policies out of the box. At test-time, the verifier can be used to perform instruction optimization and action verification, improving downstream performance for VLAs.
}
    \label{fig:teaser}
  \end{center}
}
\makeatother

\begin{document}
\maketitle
\input{sec/0_abstract}    

\input{sec/1_intro}
\input{sec/2_related_work}
\input{sec/scaling_law}
\input{sec/3_method}
\input{sec/4_Experiments}
\input{sec/5_conclusion}

% Main paper bibliography
{
    \small
    \bibliographystyle{ieeenat_fullname}
    \bibliography{bibs/others, bibs/robotics, bibs/tts, bibs/ttt}
}
\clearpage
% Note: Supplementary material is in supp_main.tex
\input{sec/suppl}

\end{document}

%% file: preamble.tex
\usepackage{booktabs}
\usepackage{xspace}
\newcommand{\abbv}{{CoVer}\xspace}

%% file: sec/0_abstract.tex
\begin{abstract}
The long-standing vision of general-purpose robots hinges on their ability to understand and act upon natural language instructions. Vision-Language-Action (VLA) models have made remarkable progress toward this goal, yet their generated actions can still misalign with the given instructions. In this paper, we investigate test-time verification as a means to shrink the ``intention-action gap.'' We first characterize the test-time scaling laws for embodied instruction following and demonstrate that jointly scaling the number of rephrased instructions and generated actions greatly increases test-time sample diversity, often recovering correct actions more efficiently than scaling each dimension independently. To capitalize on these scaling laws, we present CoVer, a contrastive verifier for vision–language–action alignment, and show that our architecture scales gracefully with additional computational resources and data. We then introduce CoVer-VLA, a hierarchical test-time verification pipeline using the trained verifier. At deployment, our framework precomputes a diverse set of rephrased instructions from a Vision-Language-Model (VLM), repeatedly generates action candidates for each instruction, and then uses the verifier to select the optimal high-level prompt and low-level action chunks. Compared to scaling policy pre-training on the same data, our verification approach yields 22\% gains in-distribution and 13\% out-of-distribution on the SIMPLER benchmark, with a further 45\% improvement in real-world experiments. On the PolaRiS benchmark, CoVer-VLA achieves 14\% gains in task progress and 9\% in success rate.
\end{abstract}
\vspace{-0.1in}

%% file: sec/1_intro.tex
\begin{figure*}[t]
    \centering
    \includegraphics[width=\textwidth]{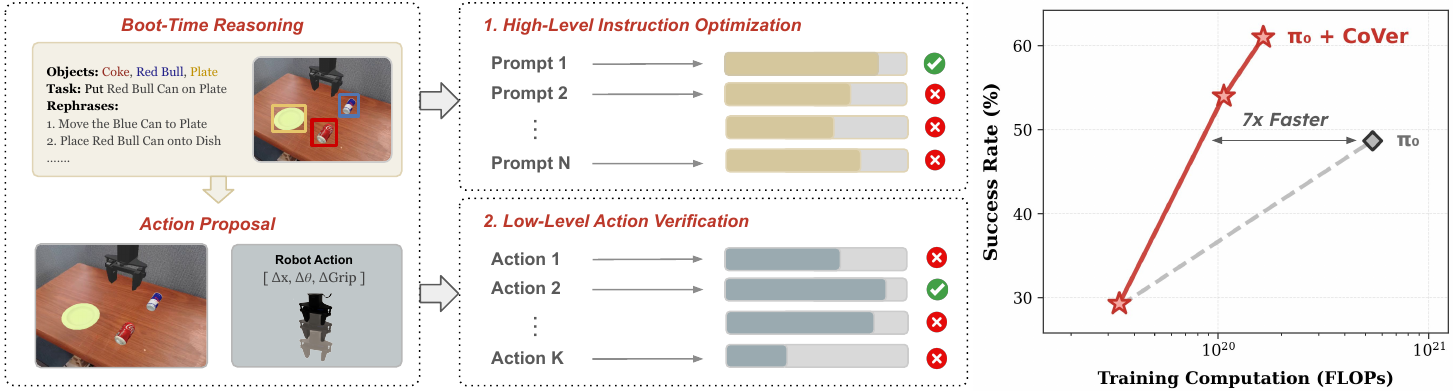}
    \caption{\textbf{Hierarchical Test-Time Verification Pipeline. } 
    \textbf{Left:} Given the initial observation and language instruction, a VLM performs structured reasoning over the scene and precomputes a set of rephrased instructions during boot time.
    At each step during deployment, our framework generates a batch of action candidates for each instruction using a VLA.
    \textbf{Middle:} CoVer then scores all instruction–action pairs and selects the optimal high-level instruction and low-level action chunk for execution.
    \textbf{Right:} Compared to prior work on scaling policy learning~\citep{black$p_0$VisionLanguageActionFlow2024}, our approach achieves stronger performance while requiring substantially less compute. The reported training compute for $\pi_0$ includes both pre-training and fine-tuning on augmented instruction sets, whereas $\pi_0$ + CoVer accounts for pre-training $\pi_0$ and training the CoVer verifier on the same data.}
    \label{fig:teaser}
    \vspace{-0.01in}
\end{figure*}

\section{Introduction}
\label{sec:intro}
For robots to be useful in human-centric environments, they must be able to interpret and act upon natural language instructions. Vision-Language-Action (VLA) models, pre-trained on large-scale robotic datasets, have made significant progress towards this goal~\citep{brohan2023rt2visionlanguageactionmodelstransfer, kim2024openvlaopensourcevisionlanguageactionmodel}. However, their widespread deployment is hindered by a critical ``intention-action gap": the misalignment between generated actions and the given language instructions. 
When the policy fails to follow the instruction, this gap can result in costly errors. For instance, a robot tasked with ``putting a plastic container into a drawer" might correctly grasp the container but then fail to discriminate between the oven and a nearby drawer, mistakenly placing the container inside the oven. The container could melt or even catch fire. Addressing this fundamental misalignment is essential for deploying robots in real-world settings.
\let\thefootnote\relax\footnotetext{† denotes Equal contribution and § indicates Equal advising.}

Existing efforts to close this gap have largely focused on scaling policy pre-training, such as augmenting training data with rephrased instructions~\citep{xu2025can} or employing larger VLM backbones~\citep{beyer2024paligemma,grattafiori2024llama}. However, these approaches typically yield only incremental gains, and performance still degrades severely under simple perturbations~\citep{fang2025intention, karnik2025embodiedredteamingauditing}. Moreover, scaling policy pre-training often leads to \textit{catastrophic forgetting}, where learning action generation diminishes the VLM’s multimodal understanding and reasoning, hindering generalization and semantic understanding~\cite{hancock2025actionslanguagefinetuningvlms, fang2025intention}. In this paper, we argue that VLA alignment can be more effectively improved through test-time scaling. More specifically, we ask in this work:

\textit{Can we enable VLAs to leverage additional computation at test time to improve the alignment between their generated actions and the provided language instructions?}

The implications of answering this question extend not only to the generalization capabilities of VLAs, but also to how practitioners should trade off pre-training and test-time compute in robotics. To this end, we first characterize the test-time scaling law for embodied instruction following. Assuming the presence of an oracle verifier, we observe that action error consistently decreases as we scale the number of rephrased instructions, establishing a clear relationship between linguistic diversity and performance gains. Moreover, we demonstrate that jointly scaling the number of rephrased instructions and the generated actions constructs a more diverse action proposal distribution. This hybrid sampling approach often recovers correct actions more efficiently than scaling each dimension independently.

To leverage these scaling laws, we seek to develop a robust verifier for both instruction optimization and action verification. Existing verifiers often focus on low-level dynamics~\cite{kwok25robomonkey, nakamoto2024steering} and require costly interactions with the environment~\citep{liu2025can}. To address this, we draw insights from cross-modal alignment~\cite{tschannen2025siglip, radford2021learning} and introduce \abbv, a \textbf{co}ntrastive approach for \textbf{ver}ifying the alignment across vision, language, and action. Our architecture employs two key components: a text-aware visual encoder that selectively extracts task-relevant features, and an action encoder that captures long-range temporal dependencies within action chunks. The results show that scaling the number of synthetic instructions, model parameters, negative samples, and verifiers in an ensemble consistently improves verification and downstream retrieval accuracy of CoVer. We train CoVer on 20 million offline samples using a 1B parameter backbone, producing a robust verifier for test-time scaling.

During deployment, our framework first leverages “boot-time compute” to let the robot reason offline. Given the initial observation and language instruction, a VLM performs structured reasoning over the scene—identifying relevant objects, spatial relations, and plausible task decompositions. The resulting reasoning traces are then used to precompute a diverse set of rephrased instructions, allowing the robot to avoid redundant rephrase generation during execution. At test time, we employ a hierarchical verification pipeline. This pipeline generates a batch of action candidates for each precomputed instruction with a VLA, scores all instruction-action pairs using CoVer, and then selects the optimal high-level instruction and low-level action chunks for execution. In summary, our contributions are as follows:
\begin{enumerate}
    \item We characterize the test-time scaling law for embodied instruction following and propose a compute-efficient action sampling method.
    
    \item We present a contrastive verifier for vision–language–action alignment and show that our architecture scales gracefully with additional computational resources and data.

    \item We introduce boot-time compute for offline embodied reasoning and a hierarchical test-time verification pipeline that couples high-level prompt optimization with low-level action chunk selection.

    \item We show that pairing VLAs with \abbv substantially improves downstream performance, achieving a 45\% absolute improvement on real-world tasks, 18\% on SIMPLER environments, and 9\% on the PolaRiS benchmark.
    
\end{enumerate}

%% file: sec/2_related_work.tex
\begin{figure*}
    \centering
    \includegraphics[width=\linewidth]{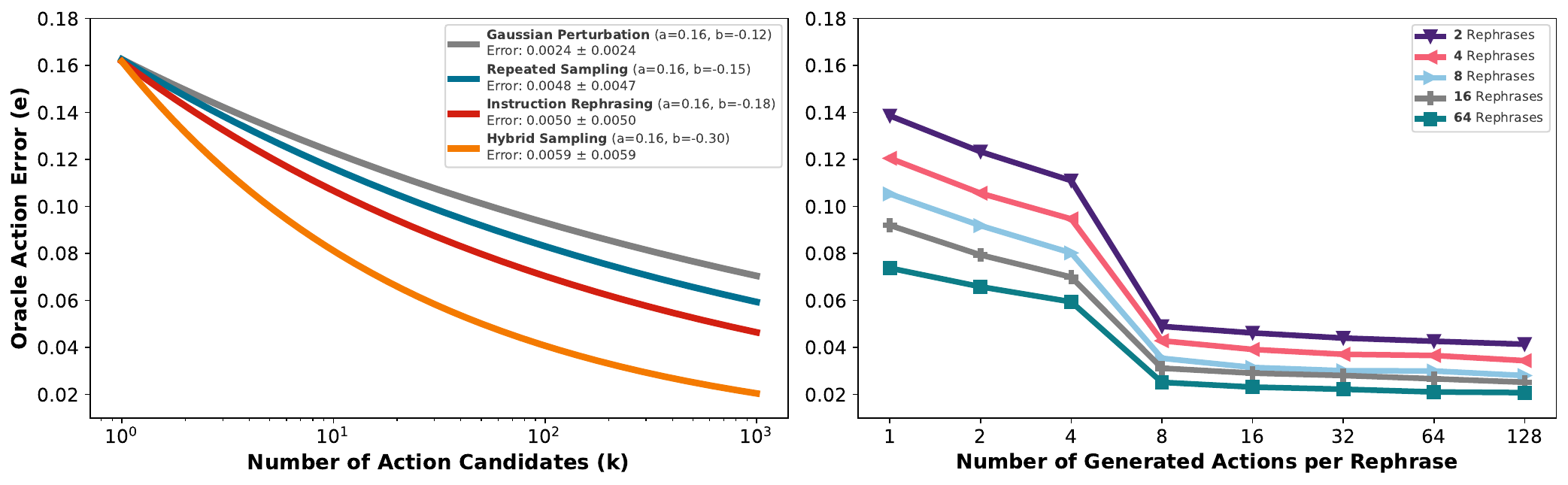}
    \caption{\textbf{Test-Time Scaling Law for Embodied Instruction Following.} 
    Compared to prior methods that construct an action proposal distribution through repeated sampling~\citep{nakamoto2024steering} or Gaussian perturbations~\citep{kwok25robomonkey}, we find that instruction rephrasing produces a broader set of action candidates, leading to improved recovery of the correct action. Furthermore, a hybrid test-time scaling strategy that increases both the number of rephrases and the number of sampled actions per rephrase is more effective than either strategy alone.
    We characterize each sampling approach using a power law, where the logarithm of oracle action error $e$ is a function of the number of action candidates $k$: $\log(e) \approx \log(a) + b \cdot \log(k)$.}
    \label{fig:scaling_law}
\end{figure*}

\section{Related Work}
\paragraph{Vision-Language-Action Models.}
Recent VLA models, pre-trained on large-scale multimodal data and fine-tuned for visuomotor control, have demonstrated impressive generalization across tasks, objects, and environments~\citep{black$p_0$VisionLanguageActionFlow2024,kim2024openvlaopensourcevisionlanguageactionmodel,nvidiaGR00TN1Open2025,teamGeminiRoboticsBringing2025,shukorSmolVLAVisionLanguageActionModel2025}.
Yet, they still struggle with instruction following: semantically equivalent rephrases can cause sharp drops in success~\citep{karnik2025embodiedredteamingauditing,fang2025intention}.
Some recent work seeks to mitigate this issue by scaling up model capacity~\citep{liGeneralistRobotPolicies2024}, expanding training data~\citep{groverEnhancingGeneralizationVisionLanguageAction2025,yangInstructVLAVisionLanguageActionInstruction2025}, and introducing auxiliary objectives to preserve linguistic knowledge~\citep{driessKnowledgeInsulatingVisionLanguageAction2025,kimContrastiveRepresentationRegularization2025}.
Orthogonal to these training approaches, our work takes a test-time perspective: we treat a user instruction as a distribution over phrasings and verify resulting actions before execution.
\vspace{-0.1in}
\paragraph{Test-Time Scaling.}
Inference with additional compute has emerged as a promising paradigm for tackling challenging problems across diverse domains, including language reasoning~\citep{snellScalingLLMTestTime2024,muennighoffS1SimpleTesttime2025,brownLargeLanguageMonkeys2024,saad-falconArchonArchitectureSearch2024}, visual understanding~\citep{wangTentFullyTestTime2020}, and agentic planning~\citep{zhangInferencetimeScalingDiffusion2025}.
In the context of robot learning, recent studies have demonstrated the effectiveness of optimizing over multiple candidate action sequences to enhance performance~\citep{nakamoto2024steering,wu2024forewarn}, consistency~\citep{liuBidirectionalDecodingImproving2025}, and robustness~\citep{kwok25robomonkey}.
Such sampling processes can be further accelerated via guidance mechanisms in the latent space~\citep{wagenmakerSteeringYourDiffusion2025,zhangAlignThenstEerAdaptingVisionLanguage2025}.
Despite these advances, existing approaches still struggle with instruction following and often incur substantial computational overhead.
Our method addresses these challenges through an action verification mechanism explicitly designed for instruction following while enabling acceleration through pre-computation.
\vspace{-0.1in}
\paragraph{Action Verification.}
Early work on action verification derives signals directly from the policy itself, e.g., prediction uncertainty~\citep{xu2025can,gu2025safe} and temporal consistency~\citep{AgiaSinhaEtAl2024,liuBidirectionalDecodingImproving2025}, yielding lightweight ways to convert prior knowledge into a quality estimator.
More recently, a growing body of work has focused on training explicit models for action verification, such as value functions~\citep{hansen-estruchIDQLImplicitQLearning2023,dongWhatMattersBatch2025} and preference models~\citep{kwok25robomonkey}.
Another line of work decomposes verification into two stages: predicting future states with a dynamics model~\citep{wu2024forewarn,qiStrengtheningGenerativeRobot2025}, and then assessing task progress in the predicted states.
However, these techniques are still largely centered on low-level dynamics, while high-level instruction following remains a challenge.
We instead formulate action verification as a contrastive alignment problem between language and behavior, explicitly targeting instruction-following quality.

%% file: sec/scaling_law.tex
\section{Test-Time Scaling Analysis}

In this section, we characterize the test-time scaling law for embodied instruction following, revealing how linguistic diversity in instructions affects downstream robot policy performance. Following the scheme introduced by Kwok et al.~\cite{kwok25robomonkey}, we uniformly sample 1,000 $(s,\ a,\ I)$ tuples from the Bridge V2 dataset~\cite{walke2023bridgedata}. For each tuple, we scale the number of generated action candidates using different sampling strategies and compute the Normalized Root Mean Squared Error (NRMSE) between the ground-truth action $a^*$ and each of the sampled actions $\{a_1,\ a_2,\ \ldots,\ a_m\}$.

We evaluate four sampling approaches: \textbf{Repeated sampling}: actions are repeatedly sampled from a robot policy $\pi(a \mid s,\ I)$ with a positive temperature. \textbf{Gaussian perturbation}: a small batch of actions is sampled from the policy $\pi(a \mid s,\ I)$, from which a Gaussian distribution is fit and used to draw all candidate actions. \textbf{Instruction rephrasing}: actions are sampled from the policy $\pi(a \mid s,\ I)$ conditioned on rephrased instructions $\{l_1,\ l_2,\ \ldots,\ l_k\}$ generated by a VLM. \textbf{Hybrid sampling}: instead of generating a single action candidate per rephrased instruction, we fan out and repeatedly sample multiple actions per rephrase. We also find that the relationship between action error and total inference FLOPs follows an exponentiated power law across these sampling methods. For power law fitting, we model the logarithm of action error $e$ as a function of the allocated inference compute.

The results in Figure~\ref{fig:scaling_law} reveal two key findings: (1) instruction rephrasing consistently yields lower action error compared to vanilla repeated sampling and Gaussian perturbation; and (2) the hybrid approach combining instruction rephrasing with repeated sampling achieves even greater diversity by exploring radically different actions rather than getting stuck in a local minimum.

%% file: sec/3_method.tex
\vspace{-0.1in}
\section{Method}
While prior works focus either on policy learning or on atomic-level action verification, our approach introduces a general hierarchical test-time verification and scaling framework (Section~\ref{sec:problem_formulation}) that integrates \textit{scalable verifier training} (Section~\ref{sec:verifier_design}) and \textit{hierarchical instruction-action verification} (Section~\ref{sec:test_time}). 
Instead of treating the base model's output as final, we jointly select high-level language prompts and low-level action alignment through an optimized latency-aware inference pipeline. 

\subsection{Hierarchical Prompt-Action Optimization}
\label{sec:problem_formulation}
We consider a sequential decision-making problem with observation space $\mathcal{O}$, action space $\mathcal{A}$, and natural-language instruction space $\mathcal{L}$.
At timestep $t$, the robot receives an observation $o_t \in \mathcal{O}$ and a user instruction $l \in \mathcal{L}$. A chunk-based VLA policy $\pi$ produces an action chunk $a_t \sim \pi(a_t \mid o_t, l)$, where $a_t$ may correspond to multiple low-level control steps.
Natural language permits many semantically equivalent rephrases, yet VLA policies
are notoriously sensitive to phrasing. 
For a rephrased instruction $l'$, the induced action $a'_t \sim \pi(a'_t \mid o_t, l')$ may deviate significantly from the intended behavior, revealing a brittleness to linguistic drift. 
This motivates treating the instruction itself as a decision variable that can be optimized at test time.
\vspace{-0.1in}
\paragraph{Language-level optimization.}
Rather than committing to a single phrasing, we construct a set with $K$ number of rephrases:
\vspace{-0.1in}
\[
\mathcal{L}_r(l') = \{\, l'_{1},\, \dots,\, l'_{K} \,\},
\]
all expressing the same user intent. 
Each $l'_k$ conditions a different action distribution under the fixed base policy. 
To formalize the objective, we use a conceptual reward function $r(o_t,a,l)$ that measures how well an action $a$ fulfills the semantics of the original instruction $l$; this reward is \emph{not} computed at test time but serves to define the ideal target behavior.
We then aim to select the rephrase whose induced behavior best aligns with the original intent:
% \vspace{-0.1in}
\[
l^*
= \arg\max_{l' \in \mathcal{L}_r}
\mathbb{E}_{a \sim \pi(\cdot \mid o_t,\, l')}
\big[ r(o_t, a, l) \big].
\]
This reformulates VLA inference as an optimization problem in \emph{language space}, not parameter space.
\vspace{-0.1in}
\paragraph{Action-level optimization.}
Given a selected rephrase $l^{*}$, sampling a single action from $\pi$ is unreliable due to bias and noise. 
We therefore draw $M$ candidate action chunks from the policy, conditioning on the current observation $o_t$ and the selected instruction $l^{*}$:
% \vspace{-0.1in}
\[
a'_{j} \sim \pi(\cdot \mid o_t,l^{*}), \qquad j=1,\dots,M,
\]

We then select the candidate that maximizes semantic alignment with the instruction:
\vspace{-0.1in}
\[
a_t^* = \arg\max_{j \in [M]}
\mathcal{V}_\theta\!\big(o_t,\, h_t,\, l^{*},\, a'_{j}\big),
\]
\vspace{-0.01in}
where $\mathcal{V}_\theta$ estimates vision–language–action alignment, and $h_t \in \mathcal{A}^{W}$ denotes the recent action history (e.g., the past $C$ actions), providing temporal context to the verifier.

This view unifies language refinement and action verification: the system first searches for the rephrase whose induced action distribution aligns with the user intent, then verifies individual action candidates within that distribution. 
Overall, developing verifier $\mathcal{V}_{\theta}$ is essential. In the next section, we will describe how to develop a \textit{robust} and \textit{scalable} verifier from available robotics datasets. 

\begin{figure}[t]
    \centering
    \includegraphics[width=\linewidth]{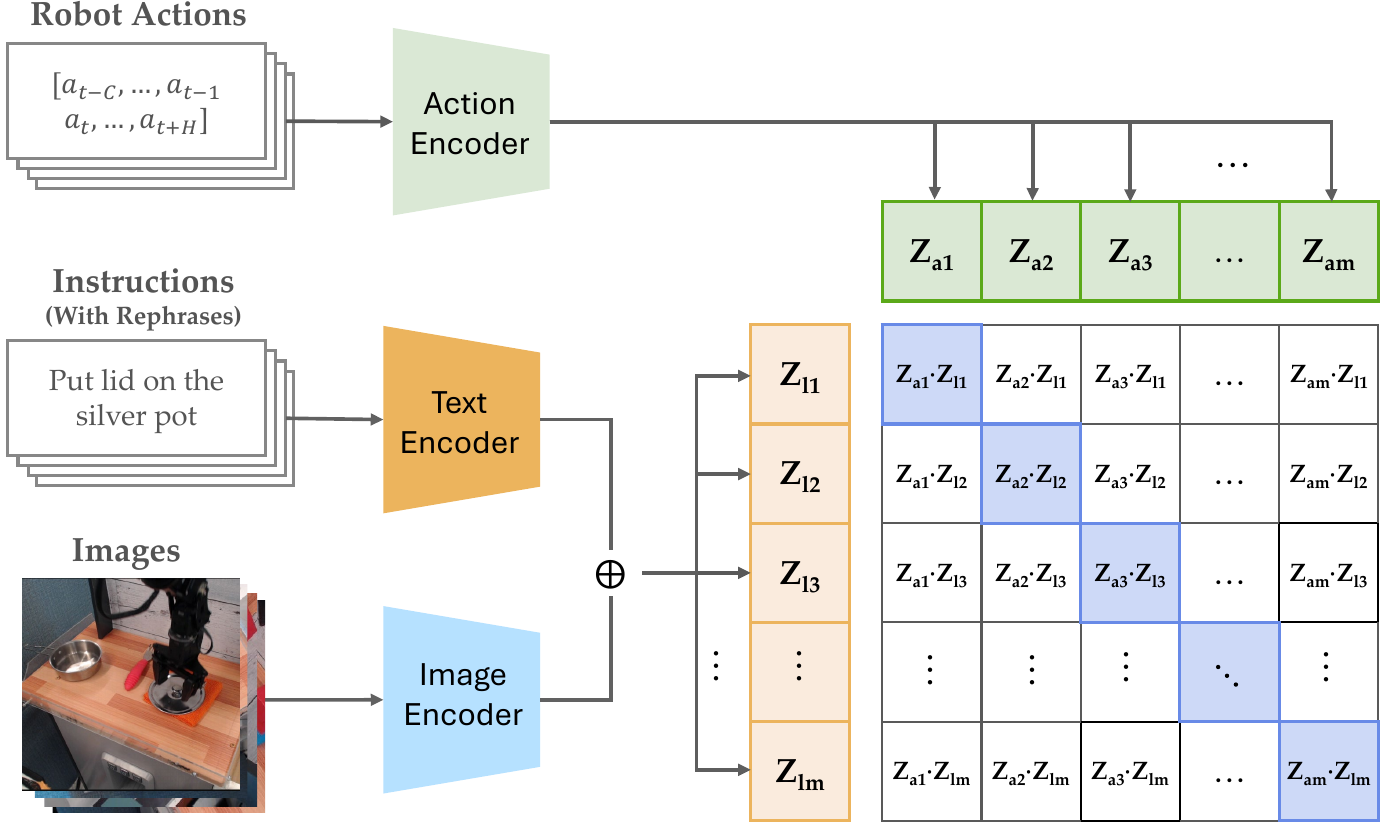}
    \caption{
    \textbf{Overview of \abbv Training Strategy.}
    \abbv learns a joint embedding space aligning visual observations, language instructions, and robot actions through contrastive pre-training. A fused image–text encoder selectively extracts task-relevant visual features, while an action encoder projects action sequences into the same embedding space. This architecture enables cross-modal alignment between high-level instructions and executed behaviors.
    }
    \label{fig:verifier_training}
\end{figure}

\subsection{Offline Verifier Training}
\label{sec:verifier_design}

The objective of verifier $\mathcal{V}_\theta$ is to assess the semantic alignment between visual observations, instruction language, and action sequences. 
A central challenge in training a VLA verifier is that robotic datasets contain only successful demonstrations, providing no direct supervision indicating when an action is semantically \emph{misaligned} with an instruction. 
Constructing negative examples is non-trivial~\citep{xu2025can}: synthesizing incorrect actions often produces unrealistic motions, while manually annotating failures is prohibitively expensive. 
Contrastive learning~\citep{radford2021learning, tschannen2025siglip} offers a natural solution by treating other actions in the batch as implicit negatives, allowing the model to learn alignment structure without curated failure labels.
Our training pipeline consists of two stages: (i) augmenting the instruction space with diverse rephrases and (ii) contrastive learning on the augmented dataset. The detailed algorithm is shown in Algorithm~\ref{alg:verifier_training}.

\begin{algorithm}[t]
\caption{Verifier Training with Rephrase Augmentation}
\label{alg:verifier_training}
\begin{algorithmic}[1]

\Require Offline trajectories $\mathcal{D}=\{(o_t,h_t,l,a_t)\}_{t=1}^T$; batch size $B$; Augmented Instruction Set $\mathcal{I}$

\State Initialize augmented dataset $\mathcal{D}_{\text{aug}} \gets \emptyset$

\Statex \textbf{Stage 1: Rephrase Augmentation}
\For{$(o_t, h_t, l, a_t) \in \mathcal{D}$}
    \For{$l^{oxe}_n$ in $\mathcal{I}(l)$}
        \State $\mathcal{D}_{\text{aug}} \gets \mathcal{D}_{\text{aug}} \cup \{(o_t, h_t, l^{oxe}_n, a_t)\}$
    \EndFor
\EndFor

\Statex \textbf{Stage 2: Verifier Training}
\State Initialize parameters $\theta$
\While{not converged}
    \State Sample minibatch $\{(o_i, h_i, l_i, a_i)\}_{i=1}^B \sim \mathcal{D}_{\text{aug}}$
    \For{$i = 1..B$}
        \State $\mathbf{F}_i = \mathbf{F}_{\text{combined}}(o_i,l_i)$ % \Comment{Vision–language embedding}
        \State $\mathbf{A}_i = \mathbf{A}(h_i, a_i)$ % \Comment{Action embedding}
        \State Normalize: $\mathbf{f}_i = \mathbf{F}_i / \|\mathbf{F}_i\|_2$, \; $\mathbf{a}_i = \mathbf{A}_i / \|\mathbf{A}_i\|_2$
        \State Compute pairwise similarities $s_{i,j} = \langle \mathbf{f}_i, \mathbf{a}_j \rangle$
        \State $\mathcal{L}_i^{f\rightarrow a} = -\log \frac{\exp(s_{i,i})}{\sum_{j=1}^B \exp(s_{i,j})}$
        \State $\mathcal{L}_i^{a\rightarrow f} = -\log \frac{\exp(s_{i,i})}{\sum_{j=1}^B \exp(s_{j,i})}$
    \EndFor
    \State $\mathcal{L}_{\text{InfoNCE}} = \frac{1}{2B} \sum_{i=1}^B (\mathcal{L}_i^{f\rightarrow a} + \mathcal{L}_i^{a\rightarrow f})$
    \State $\theta \gets \theta - \eta \nabla_\theta \mathcal{L}_{\text{InfoNCE}}$
\EndWhile

\end{algorithmic}
\end{algorithm}
\vspace{0.2em}
\vspace{-0.1in}

\paragraph{Rephrase Augmentation.}
To address the linguistic sensitivity of VLA policies, we expand each original instruction solely in language space, leaving observations and actions fixed. The training language augmentation is obtained from Open-X Embodiment~\citep{collaborationOpenXEmbodimentRobotic2023a} datasets $\mathcal{I}$, where each original task instruction set $\mathcal{I}(l)$ corresponds to $N$ rephrases.
Each selected rephrase $l^{oxe}_n$ from $\mathcal{I}(l)$ is then paired with the same observation $o_t$, and ground-truth action sequence that consists of short-term action history $h_t$ and future action chunk $a_t$ to form additional training tuples. 
Rephrases enable the verifier to encounter multiple linguistic realizations of the same underlying intent.  
This procedure enlarges the effective language coverage of the dataset without altering the action distribution, and equips the verifier with the ability to distinguish true semantic equivalence from phrasing-induced discrepancies that often mislead the base VLA policy. 
Though the same rephrase augmentation technique has been developed for policy learning~\citep{fan2025interleave, fang2025intention}, we demonstrate that using the same data budget to train a verifier would be more effective than directly augmenting the policy training dataset.

\paragraph{Verifier Training and Architecture.}
The verifier aims to estimate the alignment between visual–textual and action representations. 
Visual inputs and language tokens are encoded with pre-trained SigLIP2 encoders~\citep{tschannen2025siglip}, then fused via text-aware visual attention to obtain instruction-relevant features.
Vision and text encoders are frozen during verifier training to preserve the web-scale knowledge~\citep{huang2025otter}.
The resulting fused representation $\mathbf{F}_{\text{combined}}$ captures visual–language context. 
The action sequence, which contains short-term history and future chunks, is processed by a transformer encoder to better capture the temporal features of low-level behaviors~\citep{liuBidirectionalDecodingImproving2025}. 
The fused vision–language representation $\mathbf{F}_{\text{combined}}$ and the action embedding $\mathbf{A}$ are then $\ell_2$-normalized to get $\mathbf{f}$ and $\mathbf{a}$ respectively. 
Their similarity defines the alignment score: $s(\mathbf{f}, \mathbf{a})=\langle \mathbf{f},\,\mathbf{a} \rangle$.
Given a minibatch of $B$ tuples $\{(o_i, h_i, l_i, a_i)\}_{i=1}^B$, 
the verifier is trained with bi-directional InfoNCE~\citep{oord2018representation} objective.
% \vspace{-0.1in}
% \begin{equation}
% \small
% \begin{aligned}
% \mathcal{L}_{\mathrm{InfoNCE}}
% &= -\frac{1}{2B} \sum_{i=1}^{B}
% \Bigg[
% \log
% \frac{\exp\big(s(\hat{\mathbf{f}}_i, \hat{\mathbf{a}}_i)\big)}
% {\sum_{j=1}^{B} \exp\big(s(\hat{\mathbf{f}}_i, \hat{\mathbf{a}}_j)\big)} \\
% &\quad\;\;
% + \log
% \frac{\exp\big(s(\hat{\mathbf{a}}_i, \hat{\mathbf{f}}_i)\big)}
% {\sum_{j=1}^{B} \exp\big(s(\hat{\mathbf{a}}_i, \hat{\mathbf{f}}_j)\big)}
% \Bigg].
% \end{aligned}
% \end{equation}
This symmetrical formulation aligns vision–language embeddings 
$\mathbf{f}$ with action embeddings $\mathbf{a}$ in both directions.  
By treating all other pairs in the batch as \textit{implicit negatives}, it leverages the diversity of each minibatch to learn robust fine-grained correspondences without requiring explicit failure labels or hand-crafted counterexamples.
Such in-batch contrastive structure enables the verifier to discover meaningful distinctions between semantically aligned and misaligned behaviors, leading to more stable and cycle-consistent vision–language–action grounding during test-time verification. The verifier structure is shown in Figure~\ref{fig:verifier_training}. Given a robust verifier that can score the alignment between intentions and actions, we develop a general verification framework that can adapt to any VLA policies without additional training. 

% \vspace{-0.1in}
\subsection{Test-time Verification}
\label{sec:test_time}
\vspace{-0.05in}

\begin{figure}[t]
    \centering
    \includegraphics[width=0.97\linewidth]{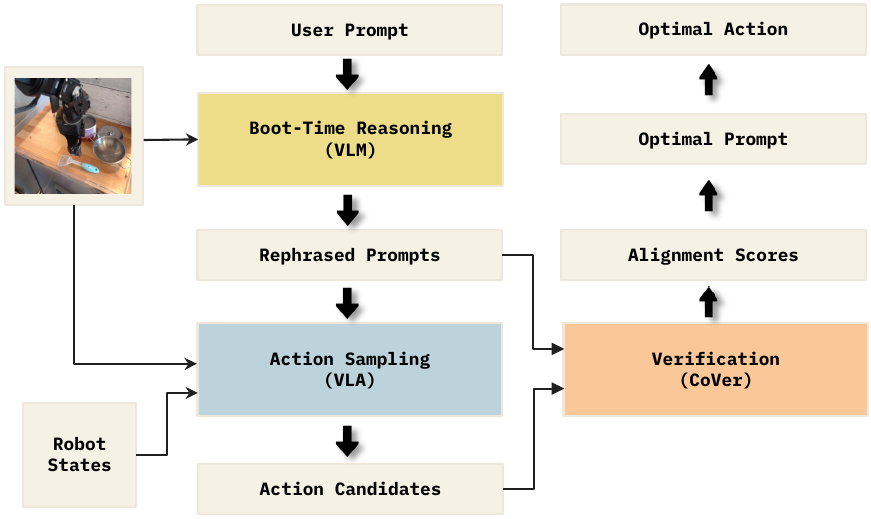}
    \caption{
    \textbf{Overview of Test-Time Verification Pipeline.}
    At deployment, the system performs hierarchical optimization over language and action spaces.
    Given a user prompt and the initial observation, a VLM first reasons over the scene and generates a set of rephrased prompts at \textit{boot time}. For each rephrase, a VLA samples action candidates conditioned on the corresponding instruction. The trained CoVer verifier then scores all instruction–action pairs and selects the optimal prompt and action for execution.
    }
    \label{fig:inference_pipeline}
\end{figure}

In Section~\ref{sec:verifier_design}, we explored the advantages of contrastive training for vision-language-action alignment, which enables zero-shot verification for both instruction and actions.
Such bidirectional features make the verification process more flexible. In this section, we propose \abbv-VLA, a test-time verification framework that is robust to language-induced action drift, while adding only minimal latency from proposal generation and verification.
\abbv-VLA casts inference as a hierarchical verification problem as shown in Figure~\ref{fig:inference_pipeline}.
The system first evaluates on the language level, selecting the instruction whose induced action distribution is most semantically reliable, and then selects the optimal action chunk conditioned on that instruction. 
This hierarchical structure enables the robot to update its active language prompt online and to filter action proposals using a learned alignment score, improving robustness without altering the underlying VLA policy.
To support this procedure, we first introduce boot-time rephrase generation and caching that significantly boosts runtime efficiency by bringing scene reasoning offline. We follow with the details on batched action proposals that enable efficient search over both languages and actions.
The resulting pipeline preserves robustness without compromising real-time control, and the full procedure is summarized in Algorithm~\ref{alg:inference}.

\vspace{-0.1in}
\begin{algorithm}[t]
\caption{Hierarchical Test-time Verification}
\label{alg:inference}
\begin{algorithmic}[1]

\State \textbf{Input:} base policy $\pi$, verifier ensemble $\mathcal{V}_\theta$, user instruction $l'$, num of rephrases $K$, num of action samples $M$
\State \textbf{Boot-time:} generate rephrases $\{l'_k\}_{k=1}^K \leftarrow \mathrm{VLM}(o_0, l')$; cache embeddings

\vspace{0.3em}
\While{episode not finished}

    \State \textbf{1. Sample action proposals}
    \For{$k = 1$ to $K$}
        \For{$j = 1$ to $M$}
            \State $a'_{k,j} \sim \pi(\cdot \mid o_t, l'_k)$
        \EndFor
    \EndFor

    \State \textbf{2. Score proposals}
    \State $s_{k,j} = \mathcal{V}_\theta(o_t, h_t, l', a'_{k,j})$

    \State \textbf{3. Select rephrase (language-level)}
    \State $S_k = \frac{1}{M} \sum_{j=1}^M s_{k,j}$ \qquad $k^* = \arg\max_{k} S_k$

    \State \textbf{4. Select action (action-level)}
    \State $j^* = \arg\max_{j} s_{k^*, j}$

    \State Execute $a'_{k^*, j^*}$ and update $(o_{t+\Delta}, h_{t+\Delta})$

\EndWhile
\vspace{-0.05in}
\end{algorithmic}
\end{algorithm}

\begin{figure*}
    \centering
    \includegraphics[width=\linewidth]{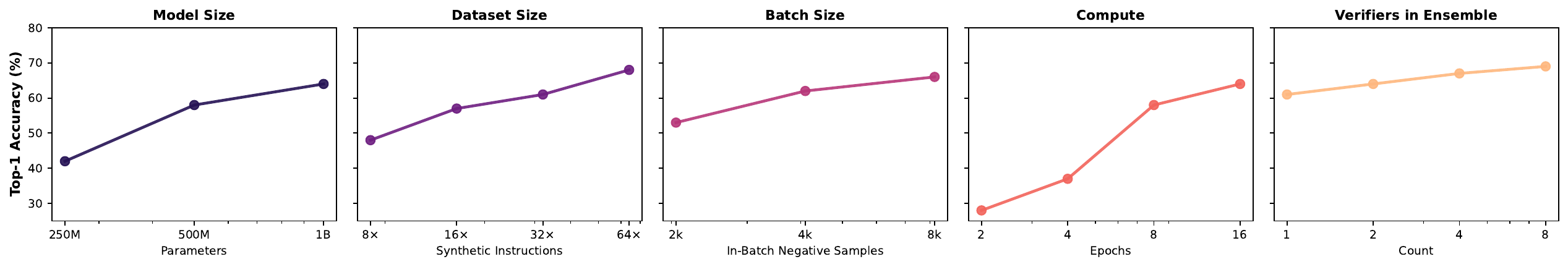}
    \caption{\textbf{Verifier Scaling Results.} We show that our architecture scales gracefully with additional compute and data. The top-1 action-retrieval accuracy consistently improves as we scale the number of synthetic instructions, model parameters, negative samples, training compute, and the number of verifiers in the ensemble. This result strongly indicates that our approach benefits from scaling, which we exploit for training CoVer.}
    \label{fig:scaling_ablation}
    \vspace{-0.1in}
\end{figure*}
\vspace{-0.1in}
\paragraph{Boot-time rephrase generation and caching.}
To efficiently handle linguistic variability, we expand each free-form instruction $l'$ into $K$ rephrases using an off-the-shelf VLM. 
The VLM takes the initial scene image $o_0$ and user instruction $l'$ as input. It generates both scene-level reasoning and rephrased command variants $\{l'_k\}_{k=1}^K$. 
Leveraging the VLM’s reasoning capabilities incorporates web-scale knowledge into the rephrase generation process.
Running the VLM on-the-fly, however, is computationally expensive and can introduce undesirable latency or motion discontinuities during robot control. 
Given that user intent is typically consistent throughout an episode, generating new rephrases mid-rollout offers limited benefits. 
Instead, we perform rephrase generation and embedding computation entirely at boot time. 
By caching rephrase embeddings before execution, we shift the heaviest computations off the critical path and ensure that retrieving rephrase features at inference time incurs negligible overhead. 
This allows the controller to evaluate paraphrastic variants efficiently at test time, enabling robust test-time optimization without compromising control smoothness. Detailed implementations of boot-time reasoning can be found in Appendix~\ref{sec:boot-time reasoning implementation}, and VLM prompts in Appendix~\ref{sec:vlm prompts for rephrase generation}.
\vspace{-0.1in}
\paragraph{Inference with batched action proposals.}
With rephrases cached and a verifier in place, we perform chunk-level optimization by jointly searching over rephrased instructions and candidate action chunks. 
Let $\{\,l'_k\,\}_{k=1}^{K}$ denote the $K$ rephrases generated at boot time, with $l'_1 = l'$.
At each chunk boundary, the base VLA policy induces a distribution over action chunks, $a \sim \pi(\cdot \mid o_t, l'_k)$, from which we sample $M$ candidates for each rephrase.  
This yields $K \times M$ proposals:
\[
a'_{k,j} \sim \pi(\cdot \mid o_t, l'_k),
\qquad k = 1,\dots,K,\;\; j = 1,\dots,M.
\]
Each proposal is then evaluated by the verifier ensemble with respect to the \textit{user instruction} $l'$,
\[
s_{k,j} = \mathcal{V}_\theta(o_t, h_t, l', a'_{k,j}),
\]
producing a semantic alignment score for every (rephrase, action) pair.
To determine which rephrase induces the most reliable action distribution,
we take the average scores across all $M$ actions from the same language:
\[
S_k = \frac{1}{M} \sum_{j=1}^M s_{k,j},
\qquad 
k^* = \arg\max_{k} S_k.
\]
The chosen rephrase $l'_{k^*}$ becomes the active language for this chunk.
Within the selected rephrase, the controller chooses the highest-scoring action candidate:
\[
j^* = \arg\max_j s_{k^*,j}.
\]
The selected action chunk $a'_{k^*, j^*}$ is executed, and the state $(o_{t+\Delta}, h_{t+\Delta})$ is updated accordingly. This procedure repeats at each chunk boundary, forming a closed-loop optimization that continually adapts both the instruction and the executed action.

%% file: sec/4_experiments.tex
\section{Experiments}
\label{experiments}

\subsection{Verifier Scaling Results}
\label{sec:main verifier scaling results}
In this section, we investigate the scaling behavior of the \abbv verifier. We conduct thorough studies to explore the impact of five key dimensions: model size, dataset size, batch size, training compute, and ensemble size. Detailed specifications regarding architecture and compute usage are provided in the Appendix.

We first evaluate how scaling synthetic instructions and model parameters affects verifier performance. As shown in Figure 4, \abbv exhibits consistent scaling trends: every increase in dataset size (from $8\times$ to $64\times$)  or in model capacity  (from 250M to 1B parameters) leads to steady improvements in top-1 retrieval accuracy. This provides strong empirical evidence that our contrastive approach effectively capitalizes on scaling. 

We also investigate the effects of scaling batch size and training epochs. Because our verifier relies on contrastive learning, the number of in-batch negative samples is critical for learning robust decision boundaries. We find that larger batch sizes (scaling from 2,048 to 8,192) provide a richer set of negative examples, thereby facilitating better convergence. Similarly, extending training epochs exposes the model to more diverse negative samples, leading to improved results. 

Finally, we explore test-time ensembling as a scaling dimension. Specifically, we train multiple verifiers with identical architectures and data budgets, differing only in their random seeds. During inference, we average the image, text, and action embeddings across these verifiers before computing the cosine similarity between modalities. We find that action retrieval accuracy consistently improves as the ensemble size increases (from 1 to 8). These gains stem from variance reduction, as the ensemble averages out individual model biases.

\subsection{Implementation Details}

Our final \abbv verifier is a 1B-parameter model trained with a batch size of 32,768 on the augmented Bridge V2 dataset~\cite{walke2023bridgedata} containing $16\times$ synthetic instructions. Training was conducted for a total of 2k steps using 8 NVIDIA H200 GPUs. For deployment, we utilize an ensemble of 3 verifiers to balance robustness and computational overhead.

\subsection{Evaluation Setup}
\label{sec:evaluation}
We evaluate \abbv-VLA across both simulated and real-world settings, focusing on robustness to linguistic variation and generalization on out-of-distribution environments (Appendix~\ref{sec:SIMPLER task description}). Our primary benchmark is the SIMPLER benchmark~\citep{li24simpler}, which includes four in-distribution (ID) manipulation tasks and three OOD variants containing distractor objects and clutter~\citep{fan2025interleave}. 
We evaluate on four representative tasks from the SIMPLER environment and adopt three challenging OOD tasks from Interleave-VLA~\citep{fan2025interleave}, includes ``Redbull on Plate", ``Zucchini on Towel", and ``Tennis in basket". 
% We run over 50 episodes on each task across 3 seeds, and report the mean success rates. 
The OOD environments contain multiple objects in the scene, where the VLA cannot rely solely on visual inputs and must also reason over the object information in the instructions. 
For real-world experiments, we use the WidowX robot to evaluate two tasks ``pepto bismol on plate'' and ``redbull on plate''. We use $\pi_{0}$ as the base model for tasks in BridgeV2. To assess how our approach performs with a stronger base policy, we also evaluate using $\pi_{0.5}$ and CoVer on the PolaRiS benchmark~\citep{jain2025polaris}. All evaluations are conducted under challenging red-teaming instructions generated by ERT~\citep{karnik2025embodiedredteamingauditing} (Appendix~\ref{sec:generated rephrases}). Our framework samples 8 rephrased instructions and generates 5 action candidates per rephrase.

\begin{figure*}
    \centering
    \includegraphics[width=\linewidth]{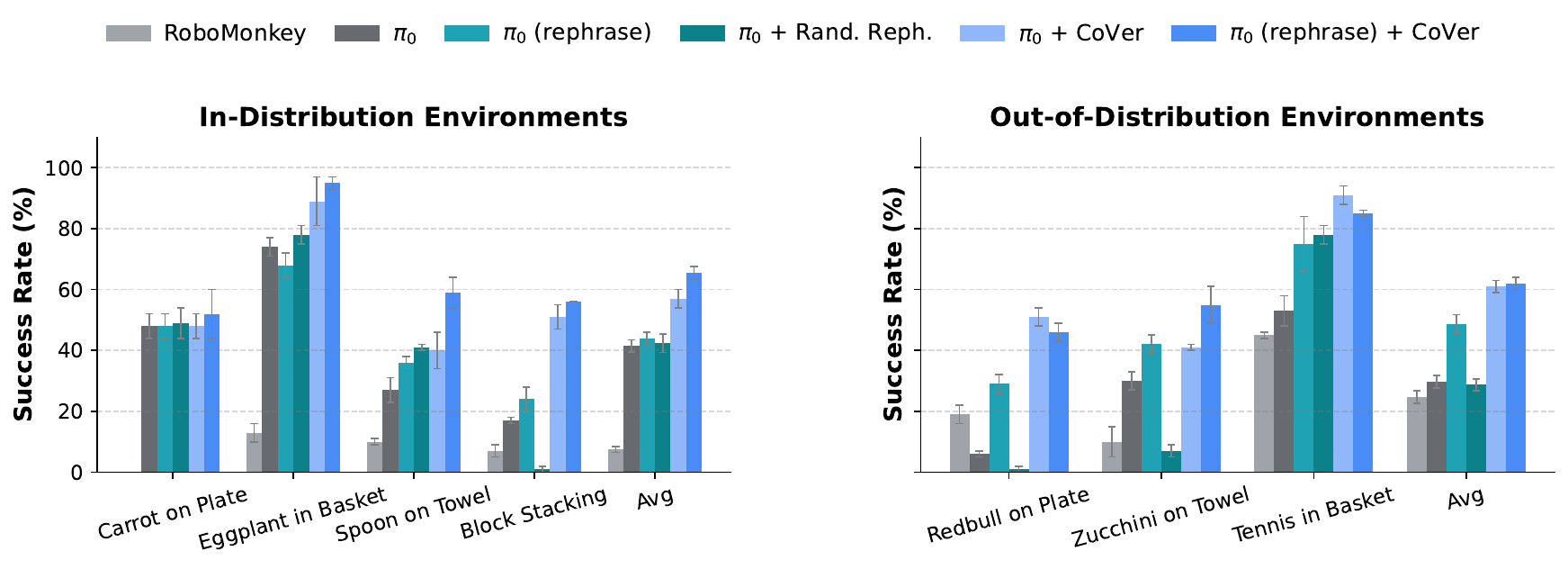}
    \caption{
    \textbf{SIMPLER Evaluation Results.} We demonstrate that scaling test-time verification with CoVer significantly enhances the robustness of VLAs across diverse manipulation tasks. Compared to scaling policy pre-training on the same data, our verification-based approach achieves a 22\% improvement on in-distribution tasks and a 13\% improvement on OOD tasks.}

    \label{fig:sim_results}
    \vspace{-0.1in}

\end{figure*}
\vspace{-0.1in}
\paragraph{Baselines and Ablations.}
We compare \abbv-VLA against five variants built on the same $\pi_0$ backbone to disentangle the effects of training-time augmentation and test-time verification (Appendix~\ref{sec:baseline and ablation descriptions}). (1)\textbf{~$\pi_0$} denotes the generalist robot policy fine-tuned on BridgeV2 without instruction augmentation or verification. (2)~\textbf{$\pi_0$ (rephrase)}~\citep{fang2025intention} represents $\pi_0$ finetuned on instruction-augmented datasets. (3)~\textbf{RoboMonkey}~\citep{kwok25robomonkey} applies a 7B-scale verifier with action resampling for test-time scaling, serving as the strongest prior method without hierarchical reasoning. (4)\textbf{$\pi_0$+~\abbv} introduces our verifier-based inference that jointly optimizes over rephrases and action chunks at test time. (5)~\textbf{$\pi_0$+Rand.~Reph.} uses a single random rephrase without verification to isolate the role of language selection. (6)~\textbf{$\pi_0$ (rephrase)+~\abbv} combines both training-time augmentation and our hierarchical test-time verifier to examine their complementarity. Together, these baselines allow us to examine the effectiveness of (i) training-time instruction augmentation, (ii) test-time verification of instructions and actions, and (iii) verifier-guided hierarchical optimization. This allows us to systematically assess \abbv’s robustness and ability to generalize across tasks.
\subsection{Simulation Evaluation Results}
\label{sec:sim_results}

Figure~\ref{fig:sim_results} summarizes performance across four ID tasks and three OOD tasks under red-teaming instructions. Detailed numerical values are described in the Appendix~\ref{sec:simulation nemerical results}.
Due to training distribution shift, Robomonkey fails to select optimal actions given challenging instructions. 
For all the other ablations, we observe different levels of performance gain over the base robot policy $\pi_0$.
We highlight three key findings below:

\textbf{(1) Training-time augmentation alone provides modest performance gain.}
We show that fine-tuning $\pi_0$ on augmented instruction sets can indeed improve robustness to challenging rephrases. However, this approach yields only minimal gains on in-distribution environments (41.5$ \rightarrow$ 44) and provides modest improvements on OOD tasks.

\textbf{(2) Random rephrases can improve performance on some tasks but lack consistency without language-level verification.}
Using a randomly generated VLM rephrase slightly improves ID performance over the base policy $\pi_0$ (41.5 $\rightarrow$ 42.3), confirming that rephrasing can enhance policy performance in some cases. However, OOD performance declines (29.7 $\rightarrow$ 28.7), and the variance across tasks is substantial. For example, the model achieves a 78\% success rate on \emph{Eggplant in Basket} but only 1\% on \emph{Redbull on Plate}. This reveals a key insight: while certain rephrased instructions can be beneficial, others may catastrophically mislead the policy. These results underscore the potential of VLM-generated rephrasings, but also expose their inconsistency.

\textbf{(3) \abbv-VLA substantially enhances generalization and complements policy learning.}
Pairing \abbv{} with $\pi_0$ significantly enhances robustness, yielding a 16\% improvement on in-distribution tasks and a 31\% gain in OOD environments. Notably, we find that scaling verification ($\pi_0$ + \abbv{}) outperforms scaling policy learning ($\pi_0$ fine-tuned with augmented instructions), achieving 15\% gains on ID tasks and 12\% on OOD, while requiring substantially less compute, as illustrated in Figure~\ref{fig:teaser}. Interestingly, our approach is complementary to scaling policy learning. Combining $\pi_0$ (rephrase) and \abbv{} achieves the strongest overall performance: 65.5\% on ID tasks and 62.0\% on OOD tasks. We further evaluate our method with a stronger base model, $\pi_{0.5}$, on the PolaRiS benchmark~\citep{jain2025polaris}. Pairing $\pi_{0.5}$ with \abbv{} leads to a 14\% improvement in task progress and a 9\% gain in success rate. By jointly selecting the semantically aligned instruction and verifying action chunks, our method reliably recovers correct behavior even under heavily perturbed instructions and in challenging OOD environments.

\begin{table}[t]{}
  \centering
  \resizebox{0.95\linewidth}{!}{
  \begin{tabular}{c|c|c|c}
    \multicolumn{1}{l|}{} & Models & Task Progress (\%) & Success Rate (\%) \\ \hline
    \multirow{2}{*}{PanClean}
      & $\pi_{0.5}$ & 48.4 $\pm$ 1.9 & 10.7 $\pm$ 0.9\\
      & $\pi_{0.5}$ + CoVer & 70.4 $\pm$ 4.0 & 33.3 $\pm$ 6.6 \\ \cline{2-4}
    \multirow{2}{*}{BlockStack}
      & $\pi_{0.5}$ & 33.1 $\pm$ 1.3 & 0.0 $\pm$ 0.0 \\
      & $\pi_{0.5}$ + CoVer & 44.3 $\pm$ 2.5 & 0.7 $\pm$ 0.9 \\ \cline{2-4}
    \multirow{2}{*}{FoodBussing}
      & $\pi_{0.5}$ & 38.3 $\pm$ 2.4 & 0.7 $\pm$ 0.9 \\
      & $\pi_{0.5}$ + CoVer & 47.0 $\pm$ 4.1 & 5.3 $\pm$ 1.9 \\ \hline
    \multirow{2}{*}{Average}
      & $\pi_{0.5}$ & 40.0 $\pm$ 6.4 & 3.8 $\pm$ 4.9 \\
      & $\pi_{0.5}$ + CoVer &
      \textbf{53.9 $\pm$ 11.7} (\textbf{+13.9$\uparrow$}) &
      \textbf{13.1 $\pm$ 14.1} (\textbf{+9.3$\uparrow$})
  \end{tabular}}
  \captionof{table}{
    \textbf{PolaRiS Evaluation Results.}
    Mean task progress and success rate ($\pm$ standard deviation) across 50 episodes and 3 seeds on three PolaRiS environments using \textbf{$\pi_{0.5}$} as the base robot policy. Pairing $\pi_{0.5}$ with CoVer consistently improves performance across all tasks, achieving a 13.9\% gain in task progress and a 9.3\% increase in success rate on average.
}
  \label{tab:average_polaris}
\end{table}

\begin{figure*}
    \centering
    \includegraphics[width=\linewidth]{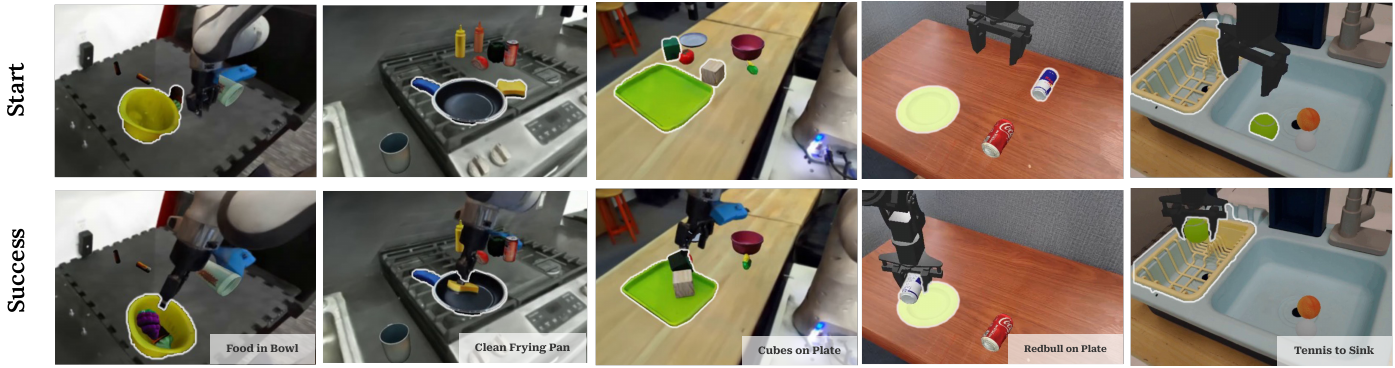}
    \caption{Example tasks across DROID~\citep{khazatskyDROIDLargeScaleInTheWild2024} and Bridge V2~\citep{walke2023bridgedata} environments.}
    \label{fig:tasksuite}
\end{figure*}
\vspace{-0.1in}

\subsection{Real-World Evaluation Results}
\label{real-world results}
We further evaluate \abbv-VLA in two real-world manipulation tasks as shown in Figure~\ref{fig:real_world results}.
\abbv-VLA substantially outperforms the baselines, improving the success rate by 30\% and 60\%, respectively. \abbv-VLA consistently shows the correct intention to accomplish the task, whereas the other baselines often fail to identify the correct object. We observe that the base $\pi_0$ model often failed to initiate motion under challenging scenes and instructions, resulting in 0\% success. Overall, these results demonstrate that scaling test-time verification with CoVer provides an effective and scalable pathway toward building a robust robotics foundation model.
% These results highlight the critical role of test-time verification in ensuring reliable deployment of robots in real-world environments.
\vspace{-0.1in}

\subsection{Latency Analysis and Optimizations}
While our approach introduces additional computational overhead from action sampling and verification, we mitigate these costs through several key optimizations. Concretely, we decouple the image-text encoder and action encoder within our verifier architecture. This design enables the image-text embedding to be computed in parallel with the forward pass of the base robot policy. As a result, the end-to-end latency of our pipeline consists only of batched inference with $\pi_0$ (or $\pi_{0.5}$) and a lightweight action encoder from CoVer. As shown in Table~\ref{tab:latency_total}, the action encoder consistently adds only $\sim$8\,ms even at larger batch sizes. In addition, repeated sampling can exploit KV cache optimizations and batch processing to achieve higher throughput than greedy decoding, allowing CoVer-VLA to sample and verify 16 candidate actions in approximately 453\,ms ($\sim$2.2\,Hz). We also avoid online rephrase generation by shifting reasoning to boot time. Specifically, we precompute and cache a set of diverse rephrased instructions before deployment. This eliminates redundant runtime calls to the VLM, thereby minimizing inference-time latency. Our full latency and throughput analysis can be found in Appendix~\ref{sec:latency and thoughput analysis}.

%% file: sec/5_conclusion.tex
\begin{figure}[t]
    \centering
    \includegraphics[width=\linewidth]{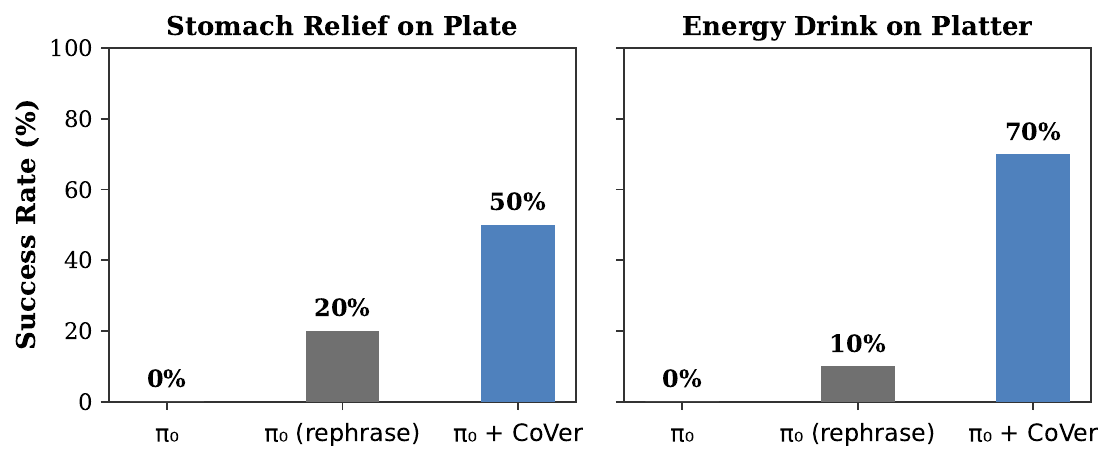}
    \caption{\textbf{Real-World Evaluation Results.} $\pi_0$ + \abbv significantly outperforms the baseline $\pi_0$ (rephrase), achieving a 45\% absolute improvement in task success rate over the baseline policy.}
    \label{fig:real_world results}
\end{figure}
\section{Conclusion}
In this paper, we present CoVer-VLA, a novel contrastive-based verifier and hierarchical test-time scaling framework that bridges the “intention–action gap” for generalist robot policies. CoVer-VLA achieves substantial performance improvements across both simulated and real-world settings, particularly under out-of-distribution conditions. Our findings demonstrate that allocating compute to reasoning and verification at deployment can be more effective than scaling policy training alone, providing a promising direction for robust policy deployment in the real world. While our study focuses on applying the verifier for test-time scaling, the same design and principle can extend beyond inference optimizations such as post-training with reinforcement learning or run-time monitoring. Future work could also explore more efficient architectures for both the base policy and verifier to further reduce latency and enable broader use of test-time scaling in real-world robotic settings. 

\begin{table}[h]
\vspace{0.16in}
\centering
\resizebox{0.9\linewidth}{!}{%
\begin{tabular}{c|ccc}
\hline
\textbf{Batch Size} &
\textbf{$\pi_{0.5}$ (ms)} &
\textbf{CoVer (ms)} &
\textbf{$\pi_{0.5}$ + CoVer (ms)} \\
\hline
1  & 56  & 7 & 63  \\
2  & 84  & 7 & 91  \\
4  & 138 & 8 & 146 \\
8  & 243 & 8 & 251 \\
16 & 445 & 8 & 453 \\
32 & 865 & 8 & 873 \\
\hline
\end{tabular}
}
\caption{Latency (milliseconds) across batch sizes running on RTX-5090 GPU. Since the image-text encoder can run in parallel with the forward pass of $\pi_{0.5}$, the end-to-end latency of our pipeline only consists of batched inference with $\pi_{0.5}$ and the lightweight CoVer action encoder}
\label{tab:latency_total}
\end{table}

\vspace{0.2in}
\section{Acknowledgments}
We thank the members of the Stanford Autonomous Systems Lab, Scaling Intelligence Lab, and IRIS Lab for their constructive feedback and informative discussions. We gratefully acknowledge the support of DARPA; NASA ULI; Schmidt Sciences; Google DeepMind; Google Research; Google Cloud; SNSF; IBM and Felicis.

%% file: sec/suppl.tex
\section{Appendix}
\begin{table*}[t]
\centering
\resizebox{\textwidth}{!}{
\begin{tabular}{lccccc|cccc}
\toprule
\textbf{Model} &
\multicolumn{5}{c|}{\textbf{In-Distribution Env}} &
\multicolumn{3}{c}{\textbf{Out-of-Distribution Env}} \\
\cmidrule(lr){2-6} \cmidrule(lr){7-10}
& Carrot on Plate & Eggplant in Basket & Spoon on Towel & Block Stacking & \textbf{Avg} 
& Redbull on Plate & Zucchini on Towel & Tennis in basket & \textbf{Avg} \\
\midrule
$\pi_{0}$ 
& 48 $\pm$ 4 & 74 $\pm$ 3 & 27 $\pm$ 4 & 17 $\pm$ 1 & \textbf{41.5}
& 6 $\pm$ 1 & 30 $\pm$ 3 & 53 $\pm$ 5 & \textbf{29.7} \\

$\pi_{0}$ w/ Inst. Aug.~\citep{fang2025intention}
& 48 $\pm$ 4 & 68 $\pm$ 4 & 36 $\pm$ 2 & 24 $\pm$ 4 & \textbf{44.0}
& 29 $\pm$ 3 & 42 $\pm$ 3 & 75 $\pm$ 9 & \textbf{48.7} \\

$\pi_{0}$ w/ random
& 49 $\pm$ 5 & 78 $\pm$ 3 & 41 $\pm$ 1 & 1 $\pm$ 1 & \textbf{42.3}
& 1 $\pm$ 1 & 7 $\pm$ 2 & 78 $\pm$ 3 & \textbf{28.7} \\

RoboMonkey~\citep{kwok25robomonkey}
&  0 $\pm$ 0 & 13 $\pm$ 3  & 10 $\pm$ 1 & 7 $\pm$ 2 & \textbf{7.5}
& 19 $\pm$ 3& 10 $\pm$ 5 & 45 $\pm$ 1 & \textbf{24.7} \\
\midrule

$\pi_0$+~\abbv
& 48 $\pm$ 4 & 89 $\pm$ 8 & 40 $\pm$ 6 & 51 $\pm$ 4 & \textbf{57.0}
& 51 $\pm$ 3 & 41 $\pm$ 1 & 91 $\pm$ 3 & \textbf{61.0} \\

$\pi_0$ (rephrase)+~\abbv
& 52 $\pm$ 8 & 95 $\pm$ 2 & 59 $\pm$ 5 & 56 $\pm$ 0 & \textbf{65.5}
& 46 $\pm$ 3 & 55 $\pm$ 6 & 85 $\pm$ 1 & \textbf{62.0} \\
\bottomrule
\end{tabular}
}
\caption{Success rates across in-distribution and out-of-distribution tasks on the SIMPLER benchmark under red-team instructions.}
\label{tab:red_team_language}
\end{table*}
\subsection{Evaluation Tasks}
\label{sec:SIMPLER task description}
As described in Section~\ref{sec:evaluation}, we evaluate our method on 7 tasks from the SIMPLER environments, 3 tasks from the PolaRiS benchmark, and 2 real-world tasks using the WidowX robot. Representative task executions for the benchmarks and real-world rollouts are shown in Figure~\ref{fig:simpler_images}. The Out-Of-Distribution (OOD) environments contains multiple distractors and several novel objects not present in BridgeV2~\citep{walke2023bridgedata}. Real-world evaluations introduce additional distribution shifts due to unavoidable differences in camera placement, workspace, lighting, and background. We provide task-specific details below.

\subsubsection{Bridge V2 Task Descriptions}

\begin{itemize}

    \item \textbf{Put Redbull Can on Plate (SIMPLER).}
    This task highlights a frequent language--vision ambiguity: the word ``red’’ appears in ``Redbull,’’ which often causes VLA policies to grasp the \emph{red} Coca-Cola can instead of the correct \emph{blue} Redbull can. The robot must therefore ground the instruction precisely and place the correct can on the plate.

    \item \textbf{Put the Zucchini on the Towel (SIMPLER).}
    This environment tests fine-grained object discrimination in OOD scenes. The robot must identify the zucchini among multiple novel objects, including a carrot. Because both are vegetables, rephrases (e.g., replacing ``zucchini’’ with ``vegetable’’) become ambiguous, making this task a direct test of whether instruction rephrasing helps when objects share semantic categories.

    \item \textbf{Put Tennis Ball into Yellow Basket (SIMPLER).}
    The sink contains a tennis ball, a ping-pong ball, and an orange. The robot must correctly identify the tennis ball in this cluttered scene and place it inside the yellow basket while ignoring the other spherical distractors.

    \item \textbf{Put Redbull Can on Plate (Real World).}
    The setup contains multiple cans with textures and color variations not present in the simulation. The robot must select the correct can and place it onto a plate despite inherent camera and lighting variation.

    \item \textbf{Put Pepto Bismol on Plate (Real World).}
    This task introduces a completely unseen object, a pepto bismol bottle and an advil bottle, whose appearance differs substantially from all training objects. The robot must ground the novel object and place it onto a plate while ignoring other distractors.

\end{itemize}
\subsubsection{PolaRiS Task Descriptions}
\label{sec: polaris task description}
\begin{figure*}[p]
    \centering
    \includegraphics[width=\textwidth,height=0.95\textheight,keepaspectratio]{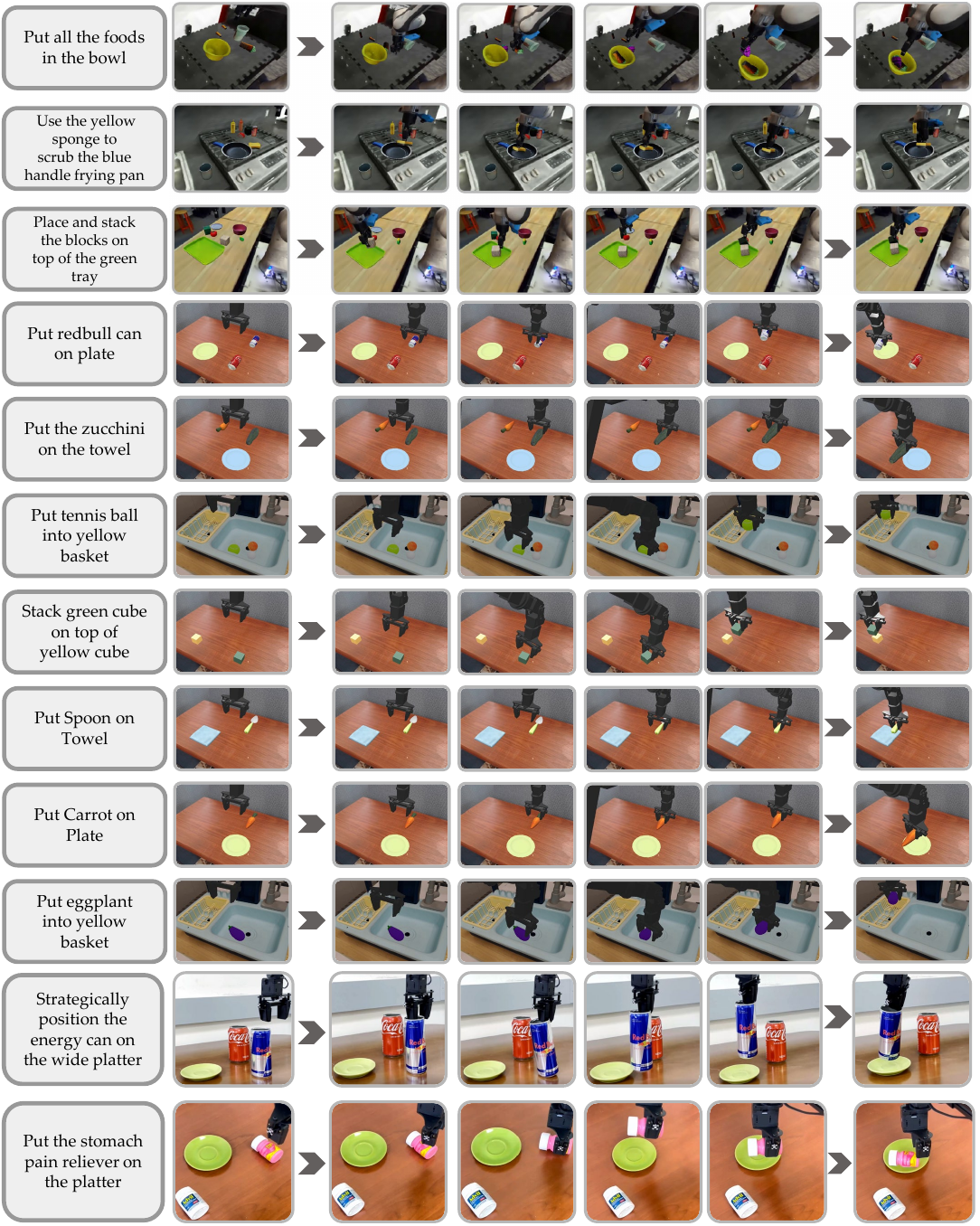}
    \caption{Task execution examples for PolaRiS, SIMPLER, and Bridge-V2 environments with corresponding original task instructions.}
    \label{fig:simpler_images}
\end{figure*}

PolaRis benchmark is developed based on DROID dataset, which contains more challenging and realistic tasks. Evaluation demonstrated on PolaRis further proved the benefits of \abbv. The successful task executions are shown in Figure~\ref{fig:simpler_images}.

\begin{itemize}
    \item \textbf{Place and stack the blocks on top of the green tray (PolaRis).}
    The table contains several distractors for both the target objects and location. The scene contains a corn, a tomato, a wooden block, a green block, a blue plate, a red bowl, and a green tray. The policy needs to accurately identify the wooden and green block, and put both object on the tray. 
    
    \item \textbf{Put all the foods in the bowl (PolaRis).}
    The scene contains two batteries, one ice cream, one grape, one cup, and one bowl. The model needs to identify the food catagery (ice cream and grape), and put them sequentially into the target container. 
    
    \item \textbf{Use the yellow sponge to scrub the blue handle frying pan (PolaRis).}
    The scene represents a standard kitchen setting with cluttered objects on the stove, including two condiment bottles, a latte cup, a sushi, a coke, a sponge, and a frying pan. The task is to pick up the yellow sponge and move it to the frying pan.
\end{itemize}

\subsection{Baselines}
\label{sec:baseline and ablation descriptions}
To make the baseline design and corresponding results explicit, we summarize each evaluated setting below:

\begin{itemize}

    \item \textbf{$\pi_0$.}  
    The base $\pi_0$ checkpoint~\citep{black$p_0$VisionLanguageActionFlow2024} fine-tuned on BridgeV2~\citep{walke2023bridgedata}.  
    This represents a vanilla generalist robot policy with \emph{no} instruction augmentation and \emph{no} test-time verification.

    \item \textbf{$\pi_0$ (rephrase)~\cite{fang2025intention}.}  
    Incorporates training-time instruction augmentation using the OpenX-Embodimenet dataset~\citep{collaborationOpenXEmbodimentRobotic2023a}.  

    \item \textbf{RoboMonkey~\citep{kwok25robomonkey}.}  
    A test-time scaling framework that uses a 7B VLM-based verifier and an action resampling strategy. For fairness, we changed RoboMonkey's base policy from OpenVLA~\citep{kim2024openvlaopensourcevisionlanguageactionmodel} to $\pi_0$.
    This baseline reflects the strongest existing test-time verification method for VLAs.

    \item \textbf{$\pi_0$ +~\abbv}
    Our verifier-driven test-time pipeline applied directly to $\pi_0$.  
    This isolates the contribution of \abbv’s hierarchical optimization—jointly selecting the most suitable rephrase and the best action chunk—without any training-time augmentation.
    We evaluate using 8 sampled rephrases and 5 repeated action samples per step. The generated 8 rephrases for each tasks are summarized in Table~\ref{tab:rephrases}.

    \item \textbf{$\pi_0$ + Rand.\ Reph.}  
    Uses a single random VLM-generated rephrase, fixed for the entire rollout and without any verification. The selected rephrases for each tasks are presented in Table~\ref{tab:rephrases}.
    This isolates the effect of \abbv's test-time language optimization: if rephrase choice mattered little, random rephrases would perform similarly to \textbf{$\pi_0$+~\abbv}.

    \item \textbf{$\pi_0$ (rephrase) +~\abbv}
    Combines training-time instruction augmentation with \abbv’s inference-time optimization.  
    This setting examines whether linguistic diversity during training and hierarchical verification at inference are complementary.

\end{itemize}

In Section~\ref{sec:sim_results}, we observe that the prior test-time verification baseline, RoboMonkey, fails catastrophically on most tasks---often performing even worse than the base $\pi_0$ model.  
We attribute this to two primary factors.  
First, RoboMonkey’s action verifier is trained on an action preference dataset derived from OpenVLA; however, the action distribution of $\pi_0$ differs substantially from OpenVLA.  
Second, due to the nature of flow-based robot policies, which generate \emph{action chunks} rather than stepwise actions, RoboMonkey’s step-level verification disrupts the  structure within each chunk and frequently selects incorrect actions, resulting in lower success rates.

\subsection{Evaluation Results}
\label{sec:simulation nemerical results}
Table~\ref{tab:red_team_language} reports the success rates for the SIMPLER benchmarks described in Section~\ref{sec:sim_results}. All evaluations are conducted using red-teaming language instructions generated by ERT~\citep{karnik2025embodiedredteamingauditing}. The corresponding task descriptions and the full set of generated rephrases are provided in Table~\ref{tab:rephrases}.

\begin{table}[h]
\centering
\begin{tabular}{l l}
\toprule
\textbf{Verifier Size} & \textbf{Backbone} \\
\midrule
250M Verifier & ViT-B/16-CLIP \\
500M Verifier & ViT-B/16-SigLIP2 \\
1B Verifier   & ViT-L/16-SigLIP2 \\
\bottomrule
\end{tabular}
\caption{Verifier model size specifications.}
\label{tab:verifier_sizes}
\end{table}

\subsection{Verifier Scaling Details}
\label{sec:verifier scalling details}

In Section~\ref{sec:main verifier scaling results}, we investigate the scaling behavior of the CoVer verifier. Below, we detail the model architectures, dataset generation pipeline, and evaluation protocols used in these studies. For our model scaling ablation, we evaluate three distinct verifier sizes, as detailed in Table~\ref{tab:verifier_sizes}. We employ pre-trained image and text encoders as the backbone for all verifiers, keeping both encoders frozen during training. Notably, we observe that increasing the size of the text encoder improves downstream verification. While the 250M and 500M variants both utilize a 90M parameter image encoder, the SigLIP2-based model leverages a $7\times$ larger text encoder (280M) compared to the CLIP text encoder (40M). This indicates that the performance gains observed in the 500M model are driven primarily by improved language representation. To construct the synthetic instruction datasets, we prompt GPT-4o to generate 128 instruction variations for each original instruction in the BridgeV2 dataset. We then embed all instructions using Qwen3-Embedding-0.6B and apply $k$-means clustering to curate rephrased subsets of varying sizes ($8\times$, $16\times$, $32\times$, and $64\times$). For evaluation, we uniformly sample 1,000 $(s, a, I)$ tuples from held-out trajectories containing unseen environments and instructions from the Bridge V2 dataset. We employ GPT-4o to generate rephrased instructions, creating a fixed action pool size of 64. We report the Top-1 Action Retrieval Accuracy. Specifically, given an observation and a task description, we evaluate how often the verifier's highest-scoring action matches the ground-truth action $a$.

\subsection{Verifier Performance Analysis}
\label{sec:appendix_verifier performance analysis}
To thoroughly evaluate CoVer's capability to select the optimal action, we conduct both quantitative and qualitative analyses.

\subsubsection{Binary Classification Performance}
We first evaluate CoVer as a binary classifier to measure its ability to discriminate between aligned (ground-truth) actions and misaligned (randomly sampled) actions. The verifier demonstrates robust discriminative performance, achieving a precision of $0.765$, a recall of $0.780$, and an F1 score of $0.772$. These results highlight CoVer’s effectiveness in identifying correct actions while reliably filtering out low-quality candidates.

\begin{figure}[h]
    \centering
    \includegraphics[width=0.95
    \linewidth]{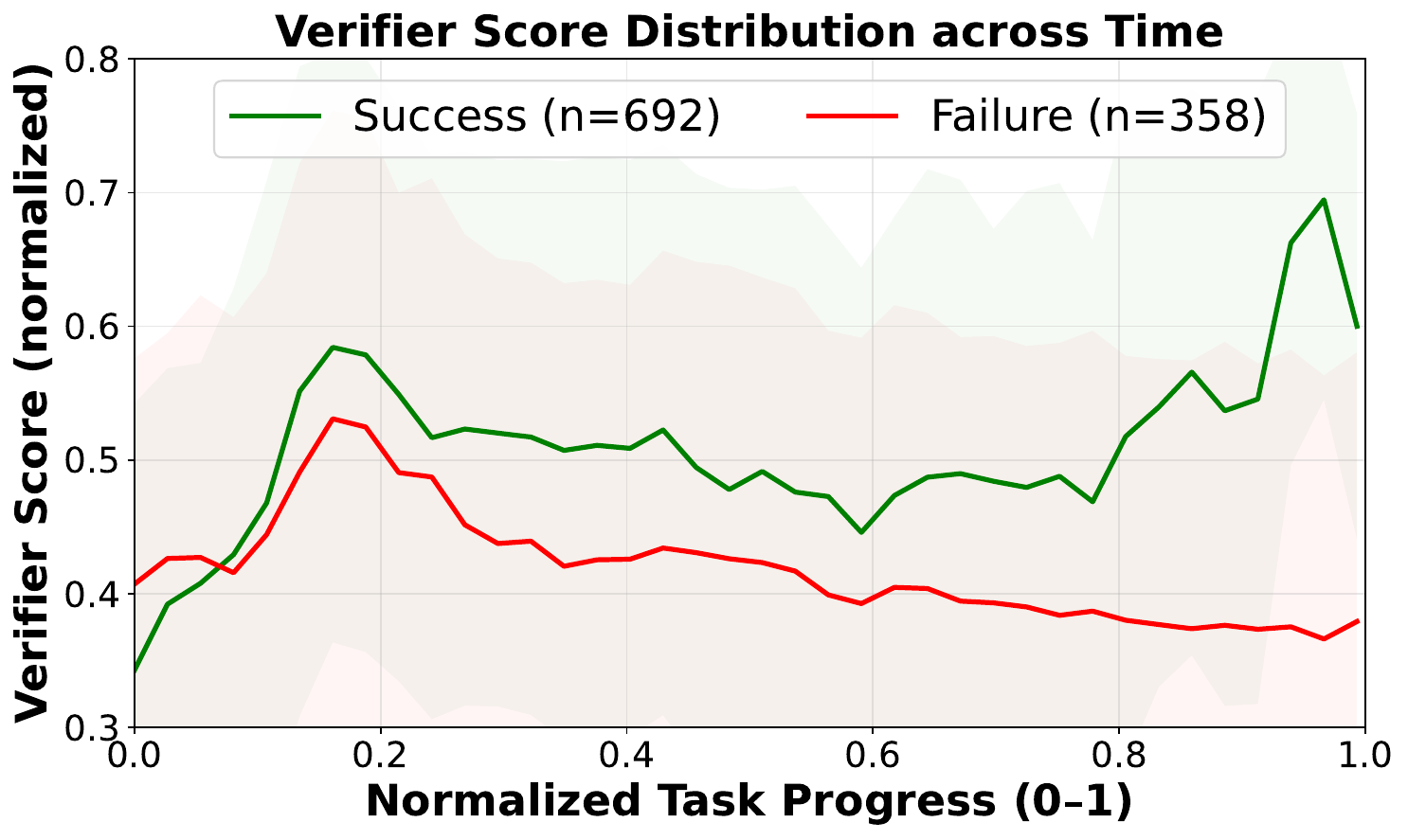}
    \caption{Visualization of verifier scores over episodes. Successful trajectories show distinct peaks during approach and completion, while failed trajectories show a steady decline.}
    \label{fig:verifier_scores}
\end{figure}
    \vspace{-8pt}

\subsubsection{Temporal Dynamics of Verifier Scores}
To visualize the verifier's behavior over the course of a rollout, we analyze the scoring distribution across episodes (see Figure~\ref{fig:verifier_scores}). We observe distinct behavioral patterns between successful and failed trajectories:
\begin{itemize}
    \item Successful trajectories consistently receive higher scores. Notably, scores peak during two critical phases: the initial approach toward the object and the final stages of task completion.
    \item Failed trajectories often exhibit a steady decline in verifier scores as the rollout progresses.
\end{itemize}
This clear separation confirms the verifier's effectiveness in identifying aligned actions and highlights its potential utility as a runtime monitor for detecting and rejecting low-confidence actions during deployment.

\begin{table*}[t]
\centering
\begin{tabular}{c|cc|cc|cc}
\hline
\multirow{2}{*}{\textbf{Batch Size}} &
\multicolumn{2}{c|}{\textbf{VLA ($\pi_0$)}} &
\multicolumn{2}{c|}{\textbf{Image+Text Encoder}} &
\multicolumn{2}{c}{\textbf{Action Encoder}} \\
\cline{2-7}
 & \textbf{Latency (ms)} & \textbf{Throughput} 
 & \textbf{Latency (ms)} & \textbf{Throughput} 
 & \textbf{Latency (ms)} & \textbf{Throughput} \\
\hline
1  & 344.0 & 2.91 & 39.0  & 25.66 & 7.0 & 145.49 \\
2  & 346.0 & 5.79 & 65.0  & 30.82 & 7.0 & 273.76 \\
4  & 374.0 & 10.71 & 79.0  & 50.82 & 8.0 & 529.01 \\
8  & 411.0 & 19.46 & 122.0 & 65.68 & 8.0 & 1060.28 \\
16 & 523.0 & 30.58 & 173.0 & 92.48 & 8.0 & 2100.90 \\
32 & 748.0 & 42.80 & 304.0 & 105.36 & 8.0 & 4204.31 \\
\hline
\end{tabular}
\caption{Latency (milliseconds) and throughput (samples/second) comparison across batch sizes}
\label{tab:latency_throughput}
\end{table*}

\subsubsection{Ablation Over Number of Samples.}
We further investigate how the number of action candidates sampled from a VLA affects the quality of the selected action. Specifically, we define action error as the RMSE between the selected action and the ground-truth action on held-out trajectories. As shown in Table~\ref{tab:candidates}, increasing the number of sampled candidates consistently reduces action error. Compared to greedy decoding ($N=1$), sampling $N=16$ candidates and selecting the optimal action via CoVer reduces the action error by $11\%$.

\begin{table}[h]
    \vspace{-8pt}
    \centering
    \small
    \begin{tabular}{cc}
    \toprule
    \textbf{\# Actions} & \textbf{RMSE} \\
    \midrule
    1  & 0.166 \\
    2  & 0.155 \\
    4  & 0.149 \\
    16 & 0.147 \\
    \bottomrule
    \end{tabular}
    \caption{Action error (RMSE) consistently decreases as we scale the number of generated action candidates.}
    \label{tab:candidates}
    \vspace{-10pt}
\end{table}

\subsection{Training Computational Cost Analysis}
\label{sec:training computational cost analysis}
To quantify the efficiency gains of our approach, we estimate the training FLOPs (floating-point operations) required for the base policy ($\pi_0$), the instruction-augmented policy ($\pi_0$ (rephrase)), and the CoVer verifier. We utilize the standard transformer training compute approximation $C \approx 6ND$, where $N$ denotes the number of parameters and $D$ represents the total number of training tokens, based on the hyperparameters provided by Fang et. al~\cite{fang2025intention}. The verifier compute is derived from the precise forward and backward pass costs per sample. Notably, because the image and text encoders are frozen during training, the backward pass does not require gradient computation for these large backbones. This results in significantly lower compute costs: the backward pass ($\approx 1.0 \times 10^9$ FLOPs) is orders of magnitude cheaper than the forward pass ($\approx 3.3 \times 10^{11}$ FLOPs).

\begin{table}[h]
    \centering
    \begin{tabular}{lcc}
        \toprule
        \textbf{Configuration} & \textbf{Total FLOPs} & \textbf{Relative Cost} \\
        \midrule
        $\pi_0$ (Base Policy) & $3.4 \times 10^{19}$ & $1.0\times$ \\
        $\pi_0$ (rephrase) ($16\times$ Data) & $5.4 \times 10^{20}$ & $16.0\times$ \\
        CoVer (Ours) & $1.3 \times 10^{20}$ & $3.8\times$ \\
        \bottomrule
    \end{tabular}
    \caption{Comparison of training computational costs.}
    \label{tab:flops_comparison}
\end{table}

\subsection{Latency and Throughput Analysis}
\label{sec:latency and thoughput analysis}
As shown in Table~\ref{tab:latency_throughput}, the $\pi_0$ batch forward pass dominates latency, rising from 344ms (batch size~1) to 748ms (batch size~32). Conversely, the \abbv action encoder incurs a constant, negligible overhead of 7--8ms. Since the image-text encoder operates in parallel with $\pi_0$, the total latency of $\pi_0$+\abbv exceeds the base model by less than 10ms in all configurations. This confirms that the verifier introduces minimal cost compared to the underlying VLA policy. 

Importantly, this small overhead has minimal impact in real-world settings. 
At the slowest tested configuration (batch size~32), the combined latency of 756\,ms corresponds to a control frequency of approximately 1.3\,Hz, which works for most quasi-static manipulation scenarios.

Overall, the measurements demonstrate that \abbv provides substantially improved robustness while preserving real-time feasibility. Note that we adopt the LeRobot implementation of $\pi_0$, available at:
\url{https://huggingface.co/juexzz/INTACT-pi0-finetune-bridge}
.

\subsection{Generated Rephrases from Red-Teaming Instructions}
\label{sec:generated rephrases}

Table~\ref{tab:rephrases} presents all instructions used across our evaluations, including original task instruction, red-team instruction, generated rephrases, and random rephrase for both SIMPLER and PolaRis.

\noindent\textbf{Original instruction.} These are template task instructions from the BridgeV2 and DROID dataset, used solely for task labeling and not included in any evaluations.

\noindent\textbf{Red-team instruction.} These challenging rephrases of bridge instructions are generated using ERT~\citep{karnik2025embodiedredteamingauditing}. We use these generated OOD instructions to evaluate the model robustness with respect to more flexible user instructions.

\noindent\textbf{Generated rephrases.} These rephrases are produced by an off-the-shelf VLM (GPT-4o) and serve as alternative instructions during the verification process. It is worth to note that the quality of generated rephrases, does not explicitly affect the verifier performance, given that the similarity score is calculated between generated actions from rephrases and the original user instruction.

\noindent\textbf{Random rephrase.} This represents a randomly selected rephrase from the generated rephrases list, used for the baseline $\pi_0$ + random rephrase evaluation.

\subsection{Boot-time Reasoning Implementation}
\label{sec:boot-time reasoning implementation}
\paragraph{Boot-time latency.}
Rephrase generation is performed once at boot time and does not incur any latency during inference, ensuring smooth execution. As such, boot-time latency is excluded in the per-step inference time reported in Table~\ref{tab:latency_total}. For reference, generating 8 rephrases with off-the-shelf VLM takes approximately 11 seconds.
\paragraph{VLM-based vs.\ LLM-based Rephrase Generation.}
Given a user instruction, we employ an off-the-shelf VLM to interpret the scene and generate instruction rephrases. 
We choose a VLM for two main reasons: (i) it provides stronger scene grounding through visual inputs, and (ii) its boot-time inference cost is negligible since it is queried only once per episode. 
Representative rephrases produced by both the VLM and a purely text-based LLM are shown in Table~\ref{tab:VLM vs LLM}. We observe that the VLM generated rephrases are generally more concise compared to LLM-based rephrases, which benefits the downstream VLA instruction understanding. For the task \textit{``put the zucchini on the towel''}, LLM-generated rephrases often include ambiguous references such as ``the vegetable,’’ which is problematic in scenes containing multiple vegetables. In contrast, the VLM reliably grounds the instruction to the correct object. Similarly, for the task \textit{``put redbull can on plate''}, the VLM produces color-specific rephrases (e.g., ``blue can’’) that significantly improve downstream VLA performance. The LLM, lacking visual grounding, instead generates category-level terms such as ``beverage,’’ which introduces semantic drift and confuses the policy.

\subsection{VLM Prompts for Rephrase Generation}
\label{sec:vlm prompts for rephrase generation}
As discussed in Section~\ref{sec:test_time}, performing rephrase generation at boot time substantially reduces inference latency by shifting both scene reasoning and linguistic diversification offline. The overall VLM prompt design follows a lightweight structure that encourages semantic preservation without imposing strong stylistic priors. 
The system prompt defines the high-level objective (rewriting manipulation instructions while keeping intent invariant), while the user prompt provides the specific instruction, the observed image, and a small set of minimally guiding examples. 
These examples serve purely as format demonstrations rather than prescriptive templates, avoiding heavy prompt engineering or over-constraining the VLM. 
In practice, this balance ensures that the model focuses on the objects and relations grounded in the scene rather than memorizing linguistic patterns from the exemplars.
To encourage accurate grounding and reduce hallucination, the user prompt explicitly asks the VLM to (i) describe the scene in its own words, (ii) reinterpret the instruction in the context of that scene, and (iii) enumerate potential lexical variations (nouns, verbs, adjectives). 
This intermediate reasoning step leads to more diverse yet semantically aligned rephrases and empirically reduces the frequency of instruction drift.
The full prompts used for generating rephrases are provided below.

\onecolumn
\paragraph{System Prompt.}
\begin{verbatim}
You are a text-transformation assistant for robot manipulation tasks.

You will be given:
- A user-provided instruction describing a manipulation goal, 
  which may involve single or multi-step actions.

Your task is to:
1. Understand the meaning of the original instruction.
2. Reword the instruction into multiple alternatives that preserve
   the original intent, and are grammatically correct and easy to follow.
3. Try to generate easy and diverse rephrases.

Guidelines:
- Reworded instructions can be diverse in terms of words, but the 
  meaning should be the same.
- Ensure all reworded instructions are semantically equivalent 
  to the original.
- Use correct grammar and clear structure.
- Keep outputs concise, consistent, and logically sound.
\end{verbatim}

\paragraph{User Prompt.}
\begin{verbatim}
Given the original instruction: "{instruction}", and the appended image, 
generate {batch_number} reworded instructions that convey the same objective.

Guidelines for rephrasing:
1. Use simple, clear words and actions (focus on verbs and nouns)
2. Remove adverbs whenever possible
3. Keep descriptions concise but complete
4. Infer and include object colors when they can be reasonably deduced 
   (e.g., apples are typically red, strawberries are red)
5. Use diverse vocabulary across rephrases (vary nouns, verbs, and adjectives)
6. Ensure each rephrase maintains the same core meaning and task objective
7. Try to generate as diverse as possible rephrases.
8. Consider the image when generating the rephrased instructions.

Examples:
Original: "put apple on the desk"
Reworded: "pick up the red apple and place it on the desk",
          "take the apple and put it on the desk",
          "place the red fruit on the desk"

Original: "put cooking pot in the green basket"
Reworded: "move the silver cooking pot to the green basket",
          "take the cooking pot and put it in the green basket",
          "put the utensil into the green basket"

Original: "put strawberry on top of the fridge"
Reworded: "put the red fruit on the fridge",
          "place the red berry on the top of the fridge",
          "set the red berry on the top of the refrigerator"

Original: "lift the water bottle and place it on the desk"
Reworded: "pick up the transparent bottle and place it on the wooden desk",
          "take the hydration bottle and put it on the desk",
          "place the water on the desk"

Guidelines for generation:
1. You need to consider both image and instruction when generating 
   the rephrased instructions.
2. You need to first generate a description of the image in your own words, 
   and then think about what the language instruction means in the 
   context of the image.

Format your response as:
<Description of the image>
<Meaning of the instruction in the context of the image>
Original: <Nouns> as many as possible potential replacements: <Nouns>
Original: <Verbs> as many as possible potential replacements: <Verbs>
Original: <Adjectives> as many as possible potential replacements: <Adjectives>
Original: <Adverbs>

Original Instruction:
{instruction}

Reworded Instructions:
1. <Alternative phrasing 1>
2. <Alternative phrasing 2>
...
{batch_number}. <Alternative phrasing {batch_number}>

Important: Ensure all rephrased instructions avoid adverbs, use 
diverse vocabulary, and maintain the same objective as the original.
\end{verbatim}

\onecolumn
\begin{longtable}{p{3.2cm} p{3.4cm} p{7.2cm} p{3.2cm}}
\caption{\textbf{Detailed description of language instructions.} The table lists the red-team instructions used as inputs to the VLA model, along with their corresponding rephrases generated by the VLM in both SIMPLER and PolaRis. We also include one randomly selected rephrase to represent the baseline \abbv + random rephrases condition.}
\label{tab:rephrases} \\
\toprule
\textbf{Original Instruction} & \textbf{Red-Team Instruction} & \textbf{Generated Rephrases} & \textbf{Random Rephrase} \\
\midrule
\endfirsthead

\multicolumn{4}{c}%
{\tablename\ \thetable\ -- \textit{Continued from previous page}} \\
\toprule
\textbf{Original Instruction} & \textbf{Red-Team Instruction} & \textbf{Generated Rephrases} & \textbf{Random Rephrase} \\
\midrule
\endhead

\midrule
\multicolumn{4}{r}{\textit{Continued on next page}} \\
\endfoot

\bottomrule
\endlastfoot

put the zucchini on the towel &
Set the green squash straight on the patterned cloth. &
1. Place the green squash upright on the blue plate. \newline
2. Put the zucchini vertically on the cloth. \newline
3. Position the squash straight on the fabric. \newline
4. Place the zucchini upright on the plate. \newline
5. Set the green vegetable vertically on the patterned fabric. \newline
6. Put the squash straight on the cloth. \newline
7. Position the zucchini on the blue cloth, standing up. \newline
8. Place the squash vertically on the plate. &
Put the squash straight on the cloth. \\
\midrule

put redbull can on plate &
Strategically position the energy can on the wide platter. &
1. Place the blue can on the yellow plate. \newline
2. Put the energy drink on the large plate. \newline
3. Set the blue drink can onto the dish. \newline
4. Lay the blue can on the wide plate. \newline
5. Position the energy drink on the yellow platter. \newline
6. Put the blue can onto the big dish. \newline
7. Place the drink can on the spacious plate. \newline
8. Set the blue can on the yellow platter. &
Put the energy drink on the large plate. \\
\midrule

stack the green block on the yellow block &
Arrange the lush green element atop the yellowish-orange element. &
1. Place the green cube on the yellow block. \newline
2. Put the grassy block on top of the yellow piece. \newline
3. Set the green block over the yellow cube. \newline
4. Stack the green piece onto the yellowish block. \newline
5. Position the lush cube above the yellow element. \newline
6. Place the green cube on the yellow cube. \newline
7. Set the green block on top of the yellow block. \newline
8. Put the green element over the yellow cube. &
Put the grassy block on top of the yellow piece. \\
\midrule

put the spoon on the towel &
Set the spoon exactly in the middle of the towel. &
1. Place the spoon in the center of the towel. \newline
2. Put the utensil on the middle of the cloth. \newline
3. Position the scoop at the center of the fabric. \newline
4. Lay the green-handled spoon in the middle of the blue towel. \newline
5. Set the utensil in the central part of the towel. \newline
6. Position the spoon on the center of the blue towel. \newline
7. Lay the green-handled spoon in the middle of the fabric. \newline
8. Set the spoon in the center of the towel. &
Put the utensil on the middle of the cloth. \\
\midrule

put carrot on plate &
Balance the carrot on the ceramic platter. &
1. Place the carrot on the green dish. \newline
2. Set the orange vegetable on the small plate. \newline
3. Position the carrot upright on the platter. \newline
4. Stand the carrot on the green plate. \newline
5. Set the vegetable on the round dish. \newline
6. Position the carrot on the small platter. \newline
7. Balance the orange carrot on the green dish. \newline
8. Place the vegetable on the ceramic plate. &
Place the carrot on the green dish. \\
\midrule

put eggplant into yellow basket &
Arrange the eggplant neatly in the yellow bin. &
1. Place the vegetable in the yellow rack. \newline
2. Put the eggplant in the yellow holder. \newline
3. Set the purple vegetable into the yellow bin. \newline
4. Organize the eggplant in the yellow rack. \newline
5. Position the vegetable in the yellow holder. \newline
6. Place the purple vegetable in the yellow container. \newline
7. Set the eggplant into the dish rack. \newline
8. Move the purple produce to the yellow bin. &
Position the vegetable in the yellow holder. \\
\midrule

put tennis ball into yellow basket &
Rest the tennis equipment in the basket that is of yellow hue. &
1. Place the tennis ball in the yellow rack. \newline
2. Set the tennis ball into the yellow container. \newline
3. Lay the tennis ball inside the lemon-colored basket. \newline
4. Put the green ball in the golden rack. \newline
5. Position the tennis equipment in the yellow dish rack. \newline
6. Rest the green ball into the yellow basket. \newline
7. Place the tennis ball inside the yellow holder. \newline
8. Lay the tennis ball in the yellow dish rack. &
Position the tennis equipment in the yellow dish rack. \\
\midrule

Put all the foods in the bowl &
Unload all foods into the bowl &
1. Put every food item in the bowl. \newline
2. Place all the food into the bowl. \newline
3. Move the edibles into the yellow bowl. \newline
4. Transfer all food objects to the bowl. \newline
5. Gather the snacks and place them in the bowl. \newline
6. Take all food pieces and put them in the bowl. \newline
7. Pick up the food and drop it into the bowl. \newline
8. Deposit all food items into the bowl. &
N/A \\
\\ \midrule

Use the yellow sponge to scrub the blue handle frying pan &
Make contact between the cleaning sponge and the frying pan with the blue handle to scrub it &
1. Scrub the blue-handled frying pan with the sponge. \newline
2. Use the yellow sponge to scrub the pan with the blue handle. \newline
3. Clean the blue-handled skillet using the sponge. \newline
4. Rub the sponge against the frying pan with the blue handle. \newline
5. Wipe the blue-handled pan with the yellow sponge. \newline
6. Use the sponge to clean the frying pan having a blue handle. \newline
7. Scrub the pan with the blue handle using the yellow sponge. \newline
8. Apply the sponge to the blue-handled frying pan to clean it. &
N/A \\
\\ \midrule

Place and stack the blocks on top of the green tray &
Carefully lay the square cubes upon the green tray &
1. Put the cubes on the green tray. \newline
2. Place the square blocks onto the green tray. \newline
3. Move the cubes to the green tray. \newline
4. Set the blocks down on the green tray. \newline
5. Position the square cubes on the green tray. \newline
6. Place the cubes into the green tray. \newline
7. Transfer the square blocks to the green tray. \newline
8. Put the square cubes on top of the green tray. &
N/A \\
\\ 

\end{longtable}

\begin{table*}[t]
\centering
\begin{tabular}{p{0.20\linewidth} | p{0.38\linewidth} | p{0.38\linewidth}}
\textbf{\centering Original Instruction} &
\textbf{VLM Rephrases} &
\textbf{LLM Rephrases} \\ \hline
\parbox[c][\height][c]{\linewidth}{\centering put the zucchini on the towel} &
\parbox[c][\height][c]{\linewidth}{
\begin{itemize}
\item Place the green squash upright on the blue plate.
\item Put the zucchini vertically on the cloth.
\item Put the squash straight on the cloth.
\item Position the zucchini on the blue cloth, standing up.
\end{itemize}
}
&
\parbox[c][\height][c]{\linewidth}{
\begin{itemize}
\item Place the green squash upright on the decorative fabric.
\item Position the vegetable vertically on the designed textile.
\item Put the green produce straight on the patterned material.
\item Arrange the squash upright on the decorative cloth.
\end{itemize}
}
\\ \hline
\parbox[c][\height][c]{\linewidth}{\centering put redbull can on plate} &
\parbox[c][\height][c]{\linewidth}{
\begin{itemize}
\item Place the blue can on the yellow plate.
\item Put the energy drink on the large plate.
\item Set the blue drink can onto the dish.
\item Lay the blue can on the wide plate.
\end{itemize}
}
&
\parbox[c][\height][c]{\linewidth}{
\begin{itemize}
\item Place the energy drink on the large tray.
\item Set the beverage can on the broad dish.
\item Arrange the energy can onto the wide tray.
\item Put the energy drink on the large dish.
\end{itemize}
}
\\ \hline
\parbox[c][\height][c]{\linewidth}{\centering stack the green block on the yellow block} &
\parbox[c][\height][c]{\linewidth}{
\begin{itemize}
\item Place the green cube on the yellow block.
\item Put the grassy block on top of the yellow piece.
\item Set the green block over the yellow cube.
\item Stack the green piece onto the yellowish block.
\end{itemize}
}
&
\parbox[c][\height][c]{\linewidth}{
\begin{itemize}
\item Place the green object on top of the orange object.
\item Set the green item over the yellow-orange piece.
\item Position the verdant piece above the amber component.
\item Put the green component on top of the gold item.
\end{itemize}
}
\\ \hline
\parbox[c][\height][c]{\linewidth}{\centering put eggplant into yellow basket} &
\parbox[c][\height][c]{\linewidth}{
\begin{itemize}
\item Place the vegetable in the yellow rack.
\item Put the eggplant in the yellow holder.
\item Set the purple vegetable into the yellow bin.
\item Organize the eggplant in the yellow rack.
\end{itemize}
}
&
\parbox[c][\height][c]{\linewidth}{
\begin{itemize}
\item Place the vegetable in an orderly fashion into the yellow container.
\item Organize the eggplant inside the lemon-colored basket.
\item Position the produce tidily in the yellow box.
\item Set the vegetable neatly in the yellow bin.
\end{itemize}
}
\\ \hline
\end{tabular}
\caption{Representative tasks where VLM and LLM rephrases differ significantly in semantics, color grounding, and linguistic drift.}
\label{tab:VLM vs LLM}
\vspace{20pt} % Adjusts space at the top
\end{table*}

\section{Notation}
\begin{table}[ht]
\centering
\begin{tabular}{ll}
\toprule
\textbf{Symbol} & \textbf{Description} \\
\midrule
$\mathcal{O}, \mathcal{A}, \mathcal{L}$ & Observation, action, and natural-language instruction spaces \\
$o_t$ & Visual observation at timestep $t$ \\
$h_t$ & Recent action history window (temporal context) \\
$l$ & Original user instruction \\
$l'_k$ & $k$-th language rephrase generated at boot-time \\
$\mathcal{L}_r(l)$ & Set of $K$ candidate rephrases \\
$\pi$ & Base Vision-Language-Action (VLA) policy \\
$a'_{k,j}$ & $j$-th action chunk sampled from $\pi$ conditioned on rephrase $l'_k$ \\
$\mathcal{V}_\theta$ & Contrastive verifier parameterized by $\theta$ \\
$s_{k,j}$ & Alignment score $\mathcal{V}_\theta(o_t, h_t, l, a'_{k,j})$ \\
$S_k$ & Average semantic reliability for the $k$-th rephrase distribution \\
$l^*$ & Selected optimal rephrase for current inference step \\
$a_t^*$ & Final verified action chunk selected for execution \\
$\mathbf{f}_i, \mathbf{a}_i$ & Normalized vision-language and action embedding vectors \\
$B$ & Training minibatch size \\
$K, M$ & Number of rephrases and action samples per rephrase \\
$\mathcal{D}_{\text{aug}}$ & Training dataset augmented with $N$ rephrases per task \\
\bottomrule
\end{tabular}
\label{tab:notation}
\end{table}